\documentclass[11pt]{article}

\usepackage[]{acl}
\usepackage{listings}
\usepackage{tcolorbox}
\usepackage{times}
\usepackage{latexsym}
\usepackage{enumitem}
\usepackage{tabularx}
\usepackage{amssymb}
\usepackage{arydshln}
\usepackage[T1]{fontenc}

\usepackage[utf8]{inputenc}

\usepackage{microtype}

\usepackage{inconsolata}

\usepackage{graphicx}

\usepackage{booktabs}
\usepackage{amsmath}
\usepackage{multirow}
\usepackage{cuted}    
\usepackage{capt-of}  

\title{Trait Collapse in Text-to-Image Generation: Evaluating and Mitigating Personality Inconsistency in Visual Storytelling}

\author{
\textbf{Yuanbo Tong, Haoyuan Li, Jibiao Jiang,} \\
\textbf{Zirui Wang, Yuchen Li, and Jiamou Liu}$^{\dagger}$ \\
School of Computer Science, The University of Auckland \\
Auckland, New Zealand \\
\texttt{jiamou.liu@auckland.ac.nz} \\
\small $^{\dagger}$Corresponding author.
}

\begin{document}
\maketitle

\begin{figure*}[t]
\centering
\includegraphics[width=0.75\textwidth]{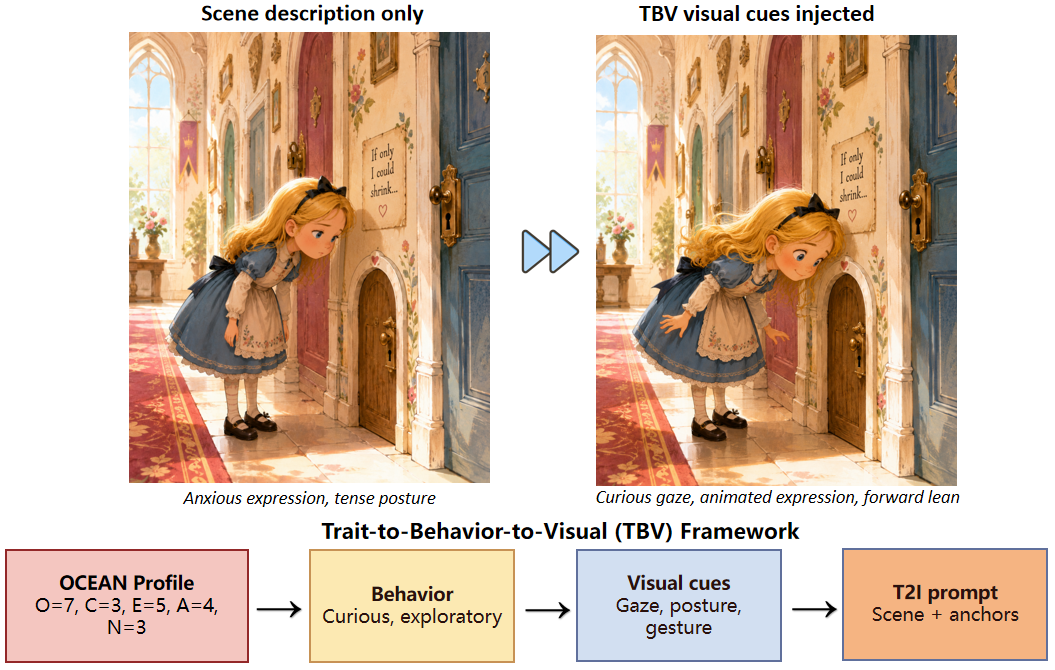}
\caption{\textbf{Trait collapse vs.\ trait expression.} Given the same narrative scene, baseline prompts (C1, left) produce generic or emotionally dominated behavior, while TBV prompts (C4, right) translate the OCEAN profile into concrete visual constraints, maintaining personality-consistent expression. Bottom: the TBV framework decomposes abstract personality traits into behavioral tendencies and then into renderable visual cues via an expert-validated lexicon.}
\label{fig:teaser}
\end{figure*}

\begin{abstract}
In visual storytelling, a believable character must not only look consistent across scenes but also \textit{behave} in ways that reflect their personality.
Current text-to-image (T2I) models preserve visual appearance effectively, yet consistently fail to maintain personality-relevant behavior---a failure we term \textit{trait collapse}.
We show that conventional metrics (CLIP Score, DINO similarity) are blind to this failure, and that abstract personality descriptors and scene-level emotion labels do not produce consistent improvements over neutral prompts across models. To systematically diagnose trait collapse, we introduce \textbf{ViStoryBench-Personality}, a story-level benchmark combining an expert-validated Visual Cues Lexicon (Fleiss' $\kappa = 0.757$) with a six-dimensional metric suite covering personality alignment, cross-scene stability, and behavioral coherence.
Evaluating seven T2I systems, we confirm a persistent semantic gap between abstract trait labels and visual evidence, and find that visual identity consistency does not imply personality consistency.
To bridge this gap, we propose \textbf{Trait-Behavior-Visual (TBV)}, a training-free prompting framework that decomposes personality traits into renderable visual constraints.
TBV improves Personality Alignment Score by up to $+22.1\%$, with larger gains in models with stronger language-conditioning capacity.
\end{abstract}

\section{Introduction}
\label{sec:intro}

In visual storytelling, character consistency extends beyond appearance.
Personality psychology treats traits as stable behavioral dispositions rather than transient emotional states \citep{mccrae1992introduction}, and observers readily infer personality from visible cues such as posture, gaze, and facial expression \citep{NaumannPersonalityAS}.
A convincing generated character should therefore not only remain visually recognizable across scenes, but also sustain behavior that is consistent with their underlying personality profile---a property we term \textit{visual personality consistency}.

Current T2I models largely fail to maintain this property.
Existing personalization methods \citep{ruiz2023dreambooth, gal2022image, ye2023ipadapter, li2024photomaker} and story-level approaches \citep{Zhou2024StoryDiffusionCS} address the question of what a character looks like, with little attention to how a character acts.
A model may faithfully preserve facial identity across frames while repeatedly producing generic smiles, rigid postures, or scene-driven emotional reactions, collapsing personality expression into a uniform behavioral default.
We term this failure \textit{trait collapse}, and attribute it to a \textit{semantic grounding} gap: models receive abstract descriptors such as ``friendly'' or ``reserved'' as input, but cannot translate them into observable, context-stable visual behavior.
This is distinct from general prompt non-compliance, as the failure is localized to personality-specific semantics rather than scene understanding.
Conventional metrics offer no signal here---CLIP score does not improve, and sometimes decreases, under personality-grounded prompting, while DINO similarity changes only marginally (mean signed change $= +0.014$; maximum $|\Delta| = 0.032$; Table~\ref{tab:clip_dino})---and existing story visualization benchmarks \citep{zhuang2025vistory} similarly do not isolate personality-relevant behavior as an evaluation target.

To address this gap, we introduce \textbf{ViStoryBench-Personality}. The benchmark represents character personality through OCEAN profiles \citep{mccrae1992introduction}, connects traits to observable visual manifestations via an expert-validated Visual Cues Lexicon ($\kappa=0.757$), and quantifies performance along six dimensions covering personality alignment, cross-scene stability, visual readability, consistency, drift, and narrative coherence.
Our evaluation shows that abstract personality descriptors and scene-level emotion labels do not produce consistent improvements over the neutral baseline across tested architectures, confirming that trait collapse is a systematic rather than incidental failure.

To mitigate trait collapse without retraining, we propose the \textbf{Trait-Behavior-Visual (TBV)} framework.
TBV decomposes personality grounding into three sequential stages: anchoring the character's OCEAN profile, deriving context-sensitive behavioral tendencies, and mapping those tendencies to concrete visual constraints drawn from the lexicon.
TBV improves Personality Alignment Score by up to $+22.1\%$, with greater gains observed in models with stronger text-encoding capacity.

\section{Related Work}
\label{sec:related}

\subsection{Character Consistency in Story Visualization}

Character preservation is a central requirement for visual narrative generation. 
T2I personalization methods preserve subject identity through test-time adaptation \citep{ruiz2023dreambooth, gal2022image} or zero-shot visual adapters \citep{ye2023ipadapter, li2024photomaker}. 
Story-level methods extend this objective to multi-shot narratives: StoryDiffusion promotes cross-frame coherence through consistent self-attention \citep{Zhou2024StoryDiffusionCS}, while recent multi-character systems introduce region-level control for identity stability \citep{banerjee2025talediffusion, cheng2025storyweaver}. 
However, these methods primarily address \textit{who} the character looks like, not \textit{how} the character behaves. 
A character may remain visually recognizable while becoming behaviorally generic, with personality-relevant posture, gaze, and interaction patterns overridden by scene-level emotion. 
This gap is consistent with findings that pretrained models often rely on broad social stereotypes rather than fine-grained trait reasoning in personality-related tasks \citep{hausladen2024social, hamidieh2024identifying, Yang2026PersonaE2AH}.

\subsection{Controllable Action and Emotion Generation}

Controllable generation methods improve control over character pose, layout, motion, and affective expression, using spatial conditions such as edges, depth, segmentation maps, or human poses \citep{zhang2023adding}, decoupled text-to-pose or motion-to-video pipelines \citep{guo2024animatediff, wang2025humandreamer}, or emotion-conditioned signals such as valence-arousal values \citep{dang2025emoticrafter}. 
However, these methods mainly specify what a character is doing or what emotion is shown in a particular scene; they do not directly evaluate whether the same character maintains stable personality expression across changing contexts. 
A model may render the requested action or emotion correctly while still producing generic, scene-driven, or personality-inconsistent behavior. 
We therefore view trait collapse as a previously under-specified failure mode: prior work has improved identity preservation, action controllability, and affective expression, but rarely examines whether characters continue to behave like the same psychological individual across scenes.

\subsection{Evaluation Benchmarks for T2I Generation}

T2I evaluation has progressed from single-image prompt adherence \citep{saharia2022photorealistic, huang2023t2i} to compositional reasoning \citep{chen2025r2ibench, deng2025psgbench} and story-level coherence \citep{zhuang2025vistory}. 
Existing story-level benchmarks assess narrative content, appearance consistency, and cross-frame identity, but do not isolate whether the same character remains psychologically coherent across scenes. 
Likewise, standard metrics such as CLIP-based text--image similarity and DINO-based appearance similarity can measure prompt alignment or visual identity, but provide no direct evidence that traits such as cautiousness, sociability, agreeableness, or emotional stability are expressed through visible behavior. 
This motivates a dedicated evaluation protocol for visual personality consistency.

\subsection{Personality Theory and Visual Behavior}

We adopt the Big Five (OCEAN) model \citep{mccrae1992introduction}, which represents stable individual differences along Openness, Conscientiousness, Extraversion, Agreeableness, and Neuroticism. 
This trait-level representation is useful because it separates enduring dispositions from transient emotions, a distinction that current T2I models often blur. 
The Brunswik Lens Model \citep{brunswik1956perception} further explains how observers infer psychological attributes from observable cues: for example, extraversion is associated with open posture, active approach, and direct gaze; agreeableness with softened expression and reduced interpersonal distance; and neuroticism with tense posture, avoidant gaze, and self-protective gestures. 
This cue-based view provides the theoretical basis for our benchmark, which evaluates personality through observable visual evidence, and for TBV, which inverts this process by translating personality profiles into concrete visual generation constraints.

\section{Problem Formulation}
\label{sec:problem}

Let $c$ denote a character with a predefined OCEAN personality profile $\mathbf{v}\in[1,7]^5$, and let $\{I_1,\ldots,I_T\}$ denote an image sequence generated for $c$ across $T$ narrative scenes.

\paragraph{Visual personality consistency.}
A generated sequence exhibits \emph{visual personality consistency} if: 
(i) each image $I_t$ contains observable personality cues---posture, gaze, facial expression, and interpersonal distance---that align directionally with $\mathbf{v}$; 
(ii) this alignment remains stable across scenes without systematic drift; and 
(iii) the alignment is legible from visual cues alone, without access to personality labels. 
This does not require the same emotion in every scene: a character may react differently across contexts while still retaining personality-consistent cues such as gaze, posture, or interaction style. 
Section~\ref{sec:metrics} operationalizes these requirements as automatic metrics.

\paragraph{Trait collapse.}
\emph{Trait collapse} occurs when visual identity consistency is maintained but visual personality consistency fails: personality-relevant cues are degraded, suppressed, or replaced by generic scene-driven reactions. 
It is distinct from identity drift, where appearance changes, and from general prompt non-compliance, where scene understanding fails broadly; the failure is localized to the abstract-to-visual mapping of personality traits. 
Section~\ref{sec:experiments} provides empirical evidence for this mechanism.

\section{ViStoryBench-Personality}
\label{sec:benchmark}

We introduce \textbf{ViStoryBench-Personality}, a diagnostic benchmark for evaluating visual personality consistency in sequential T2I generation. 
Built on ViStoryBench \citep{zhuang2025vistory}, it spans 80 stories and 1,317 sequential shots, and provides four components: structured OCEAN profiles, an expert-validated Visual Cues Lexicon, controlled prompt conditions, and a six-dimensional metric suite.

\subsection{Dataset Construction}
\label{subsec:dataset}

Each story features a focal character represented by an OCEAN vector $\mathbf{v}\in[1,7]^5$. 
From the full corpus, we sample 213 evaluation shots balanced across five affective scene categories derived from Russell's Circumplex Model \citep{russell1980circumplex}: Neutral (20.7\%), Low-arousal positive (19.2\%), Low-arousal negative (19.7\%), High-arousal positive (18.8\%), and High-arousal negative (21.6\%). 
Each shot is also labeled as \textit{Direct} (71.8\%) or \textit{Indirect} (28.2\%) depending on whether personality cues are expressed through explicit facial/gestural signals or more implicit proxemic and spatial cues. 
Full dataset statistics are provided in Appendix~\ref{app:dataset_stats}.

\paragraph{Ablation conditions.}
Each evaluation shot is generated under four prompt conditions: \textbf{C1} uses only the narrative scene and character appearance; \textbf{C2} adds abstract OCEAN trait labels; \textbf{C3} further adds a deduced scene-level emotion label; and \textbf{C4} injects explicit TBV-derived visual constraints. 
C1--C3 test whether non-visual textual personality information can reduce trait collapse, while C4 tests whether concrete visual grounding is necessary. 
A prompt example is provided in Appendix~\ref{app:prompt_example}.

\subsection{Visual Cues Lexicon}
\label{subsec:lexicon}

The Visual Cues Lexicon $\mathcal{L}_{\text{visual}}$ maps OCEAN dimensions to 150 observable cues across 10 trait poles and five signal types: facial expression, gaze behavior, body posture, clothing/color palette, and scene environment. 
Cues are assigned to three activation tiers: \textbf{Core} ($\geq5$ or $\leq3$), \textbf{Extended} ($\geq6$ or $\leq2$), and \textbf{Fine-grained} ($=7$ or $=1$), allowing visual constraint density to scale with trait extremity while avoiding over-specification for moderate profiles. 
The full lexicon is provided in Appendix~\ref{app:lexicon}.

\subsection{Expert Validation}
\label{sec:expert_validation}

We validate the Visual Cues Lexicon with 18 psychology-trained raters. 
Experts completed three tasks: OCEAN-pole mapping, trait--behavior concordance rating, and end-to-end scenario coherence judgment. 
The lexicon achieves substantial agreement in 10-way dimension mapping (Fleiss' $\kappa=0.757$), high trait--behavior concordance (S-CVI/Ave $=0.978$), and strong scenario-level coherence (S-CVI/Ave $=0.975$). 
Full task formats, per-pole statistics, and effect-size analyses are provided in Appendix~\ref{app:expert-validation}.

\subsection{Evaluation Metrics}
\label{sec:metrics}

We organize evaluation into two complementary levels: \textit{snapshot-level} metrics assess personality expression within a single image, while \textit{temporal-level} metrics assess consistency across a narrative sequence. 
All metrics use GPT-4o as the VLM judge with structured JSON outputs; the exact judge prompts and decoding parameters are provided in Appendix~\ref{app:vlm-prompts}. 
We validate this judge against human annotations (Spearman $\rho = 0.611$, polarity agreement $= 86.6\%$; Appendix~\ref{app:vlm-human}).

\subsubsection{Snapshot-Level Metrics}

\paragraph{Personality Alignment Score (PAS).}
PAS measures how faithfully a generated image reflects the intended personality. 
Given the preset OCEAN profile $\mathbf{v}$ and the VLM-inferred profile $\hat{\mathbf{v}}$, we compute centered cosine similarity as $\text{PAS}=\frac{(\hat{\mathbf{v}}-\mathbf{m})\cdot(\mathbf{v}-\mathbf{m})}{\|\hat{\mathbf{v}}-\mathbf{m}\|_2\cdot\|\mathbf{v}-\mathbf{m}\|_2}$, where $\mathbf{m}=[4,4,4,4,4]$ is the neutral midpoint. 
Centering ensures that alignment reflects directional personality evidence rather than neutral-scale bias.

\paragraph{Personality Consistency Score (PCS).}
PCS measures stochastic stability under repeated generation. 
For the same prompt, we generate $N=3$ images and compute $\text{PCS}=1-\frac{1}{5}\sum_{d\in\mathcal{D}}\frac{\mathrm{Var}(\{\hat{v}^{(i)}_d\}_{i=1}^{N})}{\sigma^2_{\max}}$, where $\sigma^2_{\max}=(U-L)^2/4=9$ by Popoviciu's inequality. 
High PCS indicates stable rendering, but not necessarily correct personality alignment, so we interpret PCS jointly with PAS.

\paragraph{Readability Score (RS).}
RS evaluates whether personality is inferable from visual cues alone. 
The VLM receives only the image and plot context, with all personality descriptors removed, and infers $\tilde{\mathbf{v}}$. 
RS uses the same centered cosine formulation as PAS between $\tilde{\mathbf{v}}$ and the target profile $\mathbf{v}$. 
High RS indicates that the generated image itself carries visually readable personality evidence.
Although RS shares the centered-cosine formulation with PAS, the two
capture related but distinct signals: across all evaluated shots they
correlate only moderately (Pearson $r=.66$), confirming that RS is not a
near-duplicate of PAS. Full inter-metric correlations are reported in
Appendix~\ref{app:metric_corr}.

\subsubsection{Temporal-Level Metrics}

\paragraph{Character Consistency Score (CCS).}
CCS measures cross-situational stability of personality expression. 
For shots $\{I_1,\ldots,I_n\}$ depicting the same character, CCS is the mean pairwise centered cosine similarity, $\text{CCS}=\frac{1}{\binom{n}{2}}\sum_{i<j}\frac{(\hat{\mathbf{v}}_i-\mathbf{m})\cdot(\hat{\mathbf{v}}_j-\mathbf{m})}{\|\hat{\mathbf{v}}_i-\mathbf{m}\|_2\cdot\|\hat{\mathbf{v}}_j-\mathbf{m}\|_2}$. 
High CCS indicates internally stable expression, but does not guarantee correctness; we therefore interpret it together with PAS and PDS.

\paragraph{Personality Drift Score (PDS).}
PDS quantifies whether personality expression systematically deviates over a narrative sequence. 
For each shot $t$, we compute the normalized distance $d_t=\frac{1}{5}\sum_{d\in\mathcal{D}}\frac{|\hat{v}^{(t)}_d-v_d|}{U-L}$ and fit $d_t=\beta_0+\beta_1\tilde{t}$ over normalized time $\tilde{t}=(t-1)/(T-1)$. 
PDS is the slope $\beta_1$: a positive value indicates progressive personality erosion, while a value near zero suggests no systematic drift.

\paragraph{Story Coherence Score (SCS).}
SCS evaluates whether the generated image aligns with both the preset personality and the behavioral logic implied by the narrative: $\text{SCS}=\alpha\cdot\text{PAS}+(1-\alpha)\cdot\text{SCS}_{\text{reasoning}}$, where $\text{SCS}_{\text{reasoning}}\in[0,1]$ is a VLM-assigned score measuring how well the image matches the personality-conditioned behavioral trace from Phase~B. We set $\alpha=0.4$ to emphasize context-specific behavioral coherence while retaining global trait alignment; a sensitivity analysis (Appendix~\ref{app:scs_sensitivity}) confirms that model rankings are stable across $\alpha \in [0.2, 0.8]$.

\section{Trait-Behavior-Visual (TBV) Framework}
\label{sec:tbv}

\begin{figure*}[t]
\centering
\includegraphics[width=0.85\textwidth]{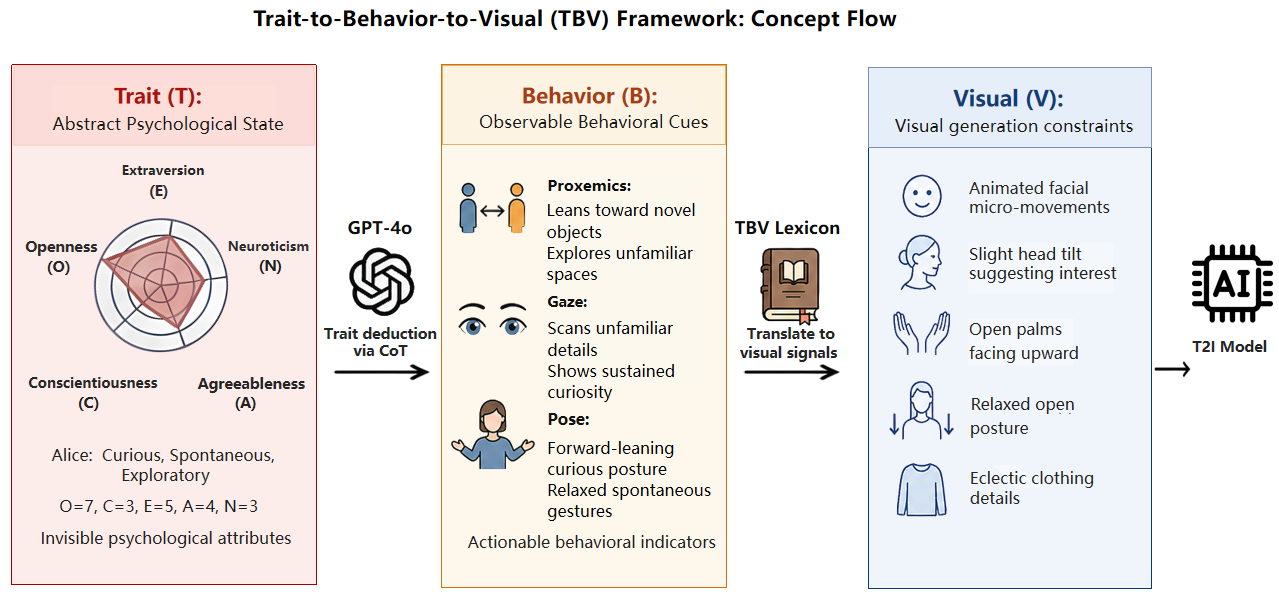}
\caption{Overview of the Trait-Behavior-Visual (TBV) framework. Phase T anchors the character's OCEAN personality profile; Phase B derives context-sensitive behavioral tendencies via a self-correcting loop; Phase V maps behaviors into concrete visual constraints through the expert-validated Visual Cues Lexicon.}
\label{fig:framework}
\end{figure*}

The ablation design of ViStoryBench-Personality (Section~\ref{subsec:dataset}) isolates whether abstract personality descriptors can reduce trait collapse by comparing conditions C1--C3. As our experiments confirm (Section~\ref{sec:experiments}), C2 and C3 do not yield consistent improvements over C1 across models, pointing to a systematic gap between abstract personality information and visual rendering. To mitigate trait collapse without model retraining, we propose the \textbf{Trait-Behavior-Visual (TBV)} framework, a training-free prompting intervention that explicitly decomposes this gap into three tractable phases. As illustrated in Figure~\ref{fig:framework}, TBV translates abstract OCEAN profiles into concrete visual generation constraints through: Phase T (Trait Anchoring), Phase B (Behavioral Deduction), and Phase V (Vision Mapping). 
Unless otherwise stated, all LLM-based operations in TBV, including trait anchoring, behavioral deduction, reverse inference, and visual-cue selection, are implemented with GPT-4o using structured JSON outputs. Implementation settings are provided in Appendix~\ref{app:implementation}.

\subsection{Phase T: Trait Anchoring}
\label{subsec:phase_t}

Phase T establishes a stable OCEAN personality anchor for each character. 
To mitigate the inherent variance of LLM-based trait extraction, we employ a multi-sample self-consistency strategy \citep{wang2023self}: we sample $k$ independent inferences and measure their agreement using a Normalized Disagreement Index, defined as $\mathrm{NDI}=\frac{1}{5k}\sum_{i=1}^{k}\sum_{d\in\mathcal{D}}\frac{|v_{i,d}-\bar{v}_{d}|}{U-L}$, where $\mathcal{D}=\{O,C,E,A,N\}$, $U=7$, $L=1$, and $\bar{v}_{d}$ is the mean inferred score for dimension $d$ across the $k$ samples. 
A three-tier routing mechanism determines the outcome: if $\mathrm{NDI}<0.15$, the mean vector $\bar{\mathbf{v}}$ is accepted as the personality anchor $\hat{\mathbf{v}}$; if $0.15 \leq \mathrm{NDI} \leq 0.33$, a multi-agent debate is triggered to reconcile divergent estimates; otherwise, the system falls back to a population-norm baseline. 
This mechanism reduces the propagation of unstable personality anchors to subsequent phases.

\subsection{Phase B: Behavioral Deduction}
\label{subsec:phase_b}

Given the anchored personality profile $\hat{\mathbf{v}}$ and a narrative scene description, Phase B generates a behavioral reasoning trace $r$ describing how the character would act in this context. 
The key challenge is ensuring that the character's response reflects their stable personality rather than defaulting to a generic reaction driven by the scene's emotional content.

To address this, we introduce a self-correcting loop inspired by iterative refinement methods \citep{madaan2023self}. 
An independent critic agent performs reverse inference on the reasoning trace to reconstruct the implied personality profile $\tilde{\mathbf{v}}=\mathrm{Inf}(r)$. 
We then compute a behavioral deviation score $s=\frac{1}{5}\sum_{d\in\mathcal{D}}\frac{|\hat{v}_{d}-\tilde{v}_{d}|}{U-L}$. 
If $s>\tau$ (we set $\tau=0.15$, equivalent to a mean deviation of 0.9 points per dimension on the 7-point scale), the trace is rejected as out-of-character and regenerated with corrective feedback. 
This mechanism ensures that the behavioral description remains personality-consistent even under emotionally charged narrative contexts.

\subsection{Phase V: Vision Mapping}
\label{subsec:phase_v}
Phase V translates behavioral descriptions into explicit visual constraints. Following an inverse application of the Brunswik Lens Model \citep{brunswik1956perception}---mapping known personality traits to the observable cues through which they become visible---we draw from the expert-validated Visual Cues Lexicon $\mathcal{L}_{\text{visual}}$ (Section~\ref{subsec:lexicon}). For each character, the lexicon selects cues matching the character's OCEAN profile at the appropriate tier level (Core, Extended, or Fine-grained).

To reduce semantic collisions (e.g., simultaneous ``clenched jaw'' from high Neuroticism and ``gentle smile'' from high Agreeableness), we apply a priority-based suppression rule: facial expression and gaze cues take precedence over posture cues, which in turn override clothing and scene-level signals. This hierarchy reflects evidence that facial behavior is the most salient channel for personality perception \citep{ryumina2026sslmepr}. The resulting conflict-reduced constraint set $\mathcal{C}_{\text{visual}}$ is injected directly into the T2I prompt as the C4 condition in our ablation design.

\section{Experiments}
\label{sec:experiments}

\subsection{Experimental Setup}
\label{subsec:exp_setup}

\paragraph{Evaluated models.}
We evaluate seven T2I systems spanning three groups: (1)~open-source generative models---\textbf{SDXL} \citep{podell2023sdxl}, \textbf{FLUX.1-dev} \citep{blackforestlabs2024flux1dev}, \textbf{SD3.5-Large} \citep{stabilityai2024sd35}, and \textbf{Qwen-Image-2512} \citep{wu2025qwenimagetechnicalreport}---covering representative UNet-, DiT-, MMDiT-, and LLM-conditioned architectures; (2)~a visual consistency method, \textbf{StoryDiffusion} \citep{Zhou2024StoryDiffusionCS}, which builds on SDXL with Consistent Self-Attention; and (3)~two closed-source APIs, \textbf{GPT Image 2} and \textbf{Gemini 3.1 Flash Image Preview (Nano Banana)}. Open-source models are evaluated in full diagnostic mode (C1--C4); closed-source APIs in benchmarking mode (C1 vs C4 only) due to internal prompt rewriting.

\paragraph{VLM judge.}
We employ GPT-4o with structured JSON output as the automated evaluator, validated against human annotations (Spearman $\rho = 0.611$, polarity agreement = 86.6\%; Appendix~\ref{app:vlm-human}) and further verified for robustness to the choice of judge in Section~\ref{sec:judge_robustness}. Detailed generation and evaluation settings are provided in Appendix~\ref{app:implementation}.

\subsection{Limitations of Conventional Metrics}
\label{subsec:conventional}

Before presenting our benchmark results, we first demonstrate that conventional T2I metrics are blind to personality dimensions. Table~\ref{tab:clip_dino} reports CLIP Score (text--image alignment) and DINO similarity (cross-frame appearance consistency) under baseline (C1) and TBV (C4) conditions across all seven models.

\begin{table}[h]
\centering
\footnotesize
\setlength{\tabcolsep}{3pt}
\begin{tabular}{@{}lcccccc@{}}
\toprule
& \multicolumn{3}{c}{\textbf{CLIP Score} $\uparrow$} & \multicolumn{3}{c}{\textbf{DINO Sim.} $\uparrow$} \\
\cmidrule(lr){2-4} \cmidrule(lr){5-7}
\textbf{Model} & C1 & C4 & $\Delta$ & C1 & C4 & $\Delta$ \\
\midrule
SDXL & .274 & .262 & $-.012$ & .443 & .475 & $+.032$ \\
FLUX.1-dev & .256 & .187 & $-.069$ & .501 & .504 & $+.003$ \\
SD3.5-Large & .260 & .252 & $-.008$ & .467 & .497 & $+.030$ \\
StoryDiff. & .259 & .256 & $-.003$ & .378 & .392 & $+.014$ \\
Qwen-2512 & .263 & .259 & $-.004$ & .644 & .631 & $-.013$ \\
GPT Image 2 & .279 & .254 & $-.025$ & .496 & .504 & $+.008$ \\
Nano Banana & .276 & .259 & $-.017$ & .487 & .509 & $+.022$ \\
\midrule
\textbf{Mean} & .267 & .247 & $-.020$ & .488 & .502 & $+.014$ \\
\bottomrule
\end{tabular}
\caption{CLIP Score and DINO similarity under C1 (baseline) and C4 (TBV). CLIP Score \textit{decreases} under TBV prompting, likely due to prompt-length effects on CLIP-based text encoders; DINO similarity changes only marginally, with a mean signed change of $+0.014$ and a maximum absolute change of $0.032$. Neither metric captures the personality alignment improvements detected by our benchmark metrics (Table~\ref{tab:main_results}).}
\label{tab:clip_dino}
\end{table}

CLIP Score shows no meaningful improvement under personality-grounded prompting---and in several cases decreases, most notably on FLUX.1-dev ($\Delta = -.069$), where TBV produces the largest PAS gains. DINO similarity changes only marginally across conditions (mean signed $\Delta = +.014$; mean $|\Delta| = .017$; maximum $|\Delta| = .032$). These results confirm that existing evaluation infrastructure cannot distinguish between personality-grounded and personality-absent generations, motivating the dedicated metrics introduced in Section~\ref{sec:metrics}.

\subsection{Main Results}
\label{subsec:main_results}

\begin{table*}[t]
\centering
\tiny
\setlength{\tabcolsep}{4pt}
\renewcommand{\arraystretch}{0.80}
\resizebox{0.80\textwidth}{!}{
\begin{tabular}{@{}lcccccc@{}}
\toprule
\textbf{Model / Setting}
& \textbf{PAS} $\uparrow$
& \textbf{PCS} $\uparrow$
& \textbf{RS} $\uparrow$
& \textbf{CCS} $\uparrow$
& \textbf{PDS} $\downarrow$
& \textbf{SCS} $\uparrow$ \\
\midrule

\multicolumn{7}{@{}l}{\textit{Open-source models: baseline}} \\
SDXL & 0.446 & 0.686 & 0.429 & 0.357 & +0.001 & 0.585 \\
StoryDiffusion & 0.480 & 0.763 & 0.467 & 0.499 & +0.000 & 0.587 \\
FLUX.1-dev & 0.464 & 0.731 & 0.469 & 0.385 & +0.003 & 0.614 \\
SD3.5-Large & 0.466 & 0.743 & 0.493 & 0.478 & +0.003 & 0.605 \\
Qwen-Image-2512 & 0.518 & 0.790 & 0.546 & 0.456 & +0.001 & 0.665 \\

\midrule
\multicolumn{7}{@{}l}{\textit{Open-source models: +TBV}} \\
SDXL + TBV & \textbf{0.467} & 0.655 & 0.423 & \textbf{0.406} & \textbf{-0.002} & 0.560 \\
StoryDiffusion + TBV & \textbf{0.502} & 0.685 & \textbf{0.508} & \textbf{0.527} & \textbf{-0.001} & 0.564 \\
FLUX.1-dev + TBV & \textbf{0.522} & \textbf{0.734} & \textbf{0.552} & \textbf{0.511} & \textbf{+0.002} & \textbf{0.642} \\
SD3.5-Large + TBV & \textbf{0.472} & \textbf{0.748} & 0.488 & 0.449 & +0.003 & 0.603 \\
Qwen-Image-2512 + TBV & \textbf{0.633} & \textbf{0.816} & \textbf{0.641} & \textbf{0.558} & \textbf{-0.000} & \textbf{0.701} \\

\midrule
\multicolumn{7}{@{}l}{\textit{Closed-source models: baseline}} \\
GPT Image 2 & 0.538 & 0.793 & 0.556 & 0.504 & -0.002 & 0.702 \\
Nano Banana & 0.542 & 0.754 & 0.551 & 0.459 & +0.002 & 0.707 \\

\midrule
\multicolumn{7}{@{}l}{\textit{Closed-source models: +TBV}} \\
GPT Image 2 + TBV & \textbf{0.648} & \textbf{0.815} & \textbf{0.659} & \textbf{0.606} & \textbf{-0.003} & \textbf{0.740} \\
Nano Banana + TBV & \textbf{0.638} & 0.748 & \textbf{0.661} & \textbf{0.505} & +0.002 & \textbf{0.730} \\

\bottomrule
\end{tabular}
}
\caption{Personality consistency evaluation across seven T2I models. TBV yields up to $+22.1\%$ relative PAS improvement on Qwen-Image-2512, with larger gains observed in models equipped with stronger text encoders.}
\label{tab:main_results}
\end{table*}

Table~\ref{tab:main_results} reports evaluation results across six metrics. 
Qualitative case studies in Appendix~\ref{app:per-model-analysis} further illustrate this pattern using the same representative story shot across models and prompting conditions (Figures~\ref{fig:qual_open} and~\ref{fig:qual_closed}). 
Baseline models show consistently low personality alignment, with PAS ranging from 0.446 (SDXL) to 0.542 (Nano Banana). 
Applying TBV increases PAS numerically across all seven systems, with statistically significant gains concentrated in models with stronger text encoders, as further analyzed in Section~\ref{subsec:ablation}. 
Qwen-Image-2512 (LLM-conditioned) and GPT Image 2 both exceed +20\% relative PAS improvement, while SDXL (CLIP-only encoder) shows the smallest gain.

Notably, StoryDiffusion achieves the strongest baseline PCS (0.763) but only moderate PAS (0.480), confirming that visual identity consistency does not imply psychological consistency. After TBV, its PAS improves to 0.502, but the gain (+4.6\%) is smaller than models with stronger language understanding, suggesting that personality grounding is bottlenecked by the text encoder rather than the visual consistency mechanism.

A related pattern emerges in PCS: TBV reduces PCS in two of the five
open-source systems (SDXL: $0.686 \rightarrow 0.655$; StoryDiffusion:
$0.763 \rightarrow 0.685$), and in three of the seven systems overall
when the closed-source Nano Banana is included ($0.754 \rightarrow
0.748$). One possible interpretation is that baseline PCS can be
inflated by generic behavioral defaults that remain similar across
seeds, whereas TBV introduces more personality-specific visual
variation. Because our experiments do not directly identify the causal
source of this PCS reduction, we treat this explanation as a hypothesis
rather than a confirmed mechanism. PCS and PAS should therefore be
interpreted jointly: a high-PCS/low-PAS profile may indicate stable but
weakly personality-aligned generation, whereas a moderate-PCS/high-PAS
profile may indicate stronger personality expression with greater
stochastic variation.

We note that PDS remains near zero across all models and conditions
(all $|\text{PDS}| \le 0.003$), indicating no systematic personality
drift over the evaluated sequences. Because our evaluation subset
samples relatively few shots per story, the temporal span may be
insufficient to expose gradual drift; we therefore retain PDS in
Table~2 for completeness but treat it as exploratory, noting it may
become more informative for longer narrative sequences.

\subsection{Robustness to Judge Choice}
\label{sec:judge_robustness}

Because all personality metrics rely on GPT-4o, a natural concern is
whether the observed TBV gains reflect genuine improvements in
personality expression or an artifact of the judge sharing priors with
the cue-to-trait mappings used in TBV construction. To test this, we re-evaluated PAS and RS using \textbf{Gemini~2.5~Flash}, a judge from a different model family that played no role in constructing TBV, the Visual Cues Lexicon, or the benchmark. For each of SDXL, FLUX.1-dev, and Qwen-Image-2512, we sampled 80 matched story shots generated under both C1 and C4, yielding 80 paired observations per model.

As shown in Table~\ref{tab:judge_robustness}, TBV gains replicate in
direction on all three models and remain statistically significant on
Qwen-Image-2512 ($p=.021$) and FLUX.1-dev ($p=.040$; paired one-sided Wilcoxon signed-rank test, testing the pre-specified directional hypothesis that TBV improves PAS over C1), preserving the ordering in which the
largest gain occurs on the LLM-conditioned model. Inter-judge
agreement on per-shot PAS (Pearson $r=.60$) is comparable to
our VLM--human agreement ($\rho=.61$; Appendix~\ref{app:vlm-human}),
indicating that $\sim$0.6 reflects the intrinsic difficulty of
single-image personality inference rather than judge-specific bias.
Under this independent judge, the Readability Score---which withholds
all personality information from the judge---shows even larger gains
than PAS ($+0.11$ on Qwen-Image-2512, $+0.10$ on FLUX.1-dev), which is
inconsistent with a keyword-recognition explanation. Full per-model
statistics are provided in Appendix~\ref{app:judge_robustness}.

\begin{table}[t]
\centering
\small
\setlength{\tabcolsep}{5pt}
\begin{tabular}{lcccc}
\toprule
\textbf{Model} & \textbf{C1} & \textbf{C4} & \textbf{$\Delta$} & \textbf{$p$} \\
\midrule
Qwen-Image-2512 & 0.534 & 0.607 & $+0.073$ & $.021$ \\
FLUX.1-dev      & 0.540 & 0.599 & $+0.059$ & $.040$ \\
SDXL            & 0.426 & 0.481 & $+0.065$ & n.s. \\
\bottomrule
\end{tabular}
\caption{TBV gains (C4 vs.\ C1 PAS) under an independent judge
(Gemini~2.5~Flash). Results are based on 80 matched C1--C4 story-shot pairs per model. Gains replicate in
direction across all three models, with the largest gain preserved on
Qwen-Image-2512. $p$ from paired one-sided Wilcoxon signed-rank tests;
n.s.\ = not significant.}
\label{tab:judge_robustness}
\end{table}

\subsection{Ablation Study}
\label{subsec:ablation}
We further examine whether trait collapse can be mitigated by adding non-visual textual information alone.
Across the five open-source models, abstract OCEAN descriptors (C2) and scene-level emotion labels (C3) produce inconsistent effects: FLUX.1-dev slightly improves under C2 (+0.019), whereas SDXL decreases by 0.054.
Paired Wilcoxon tests and TOST equivalence tests further support this pattern: 9 of 10 C1\,vs.\,C2/C3 comparisons show no significant PAS improvement, and 6 of 10 fall within a practically negligible margin ($\delta=0.05$; Appendix~\ref{app:stat_tests}).
In contrast, C4 yields significant PAS gains for models with stronger text encoders, including FLUX.1-dev, Qwen-Image-2512, GPT Image 2, and Nano Banana.
These results indicate that concrete visual grounding, rather than richer abstract labels, is the key factor for mitigating trait collapse, with efficacy modulated by text-encoding capacity.
Full ablation results are provided in Appendix~\ref{app:ablation}.

\subsection{Emotion Override Effect}
\label{subsec:emotion_override}

We further examine whether affective scene context interferes with personality expression. 
Across valid generated images, PAS differs significantly across affective scene categories (ANOVA: $F(4,5076)=20.06$, $p<.001$; Kruskal--Wallis: $H=82.00$, $p<.001$; Appendix~\ref{app:emotion_override}). 
Neutral scenes obtain the highest mean PAS, while all emotionally salient categories show lower PAS, suggesting that current T2I models tend to prioritize transient scene emotion over stable personality cues when the two signals compete. 
For example, a highly open character may be rendered primarily as anxious or withdrawn in a tense scene, even when the intended personality profile implies exploratory gaze or forward-leaning engagement. 
This finding supports our central claim that trait collapse is not merely an identity-preservation problem, but a semantic grounding problem: personality must remain visually legible even when emotional context changes. 
We also test whether personality alignment differs between direct and indirect visual-expression shots, but find no significant gap; this null result is reported in Appendix~\ref{app:indirect}.

\section{Discussion and Conclusion}
\label{sec:discussion}

This paper identifies and formalizes \textit{trait collapse}, a failure mode in which T2I models preserve visual identity while losing personality-relevant behavior. 
Through ViStoryBench-Personality, we show that this failure is systematic across seven T2I systems: abstract OCEAN labels and scene-level emotion labels do not consistently improve personality alignment across models, while TBV yields numerical PAS gains across all systems, with significant gains concentrated in stronger language-conditioned models. 
These results suggest that the bottleneck is not simply that models lack personality words in the prompt, but that they fail to ground abstract psychological semantics into renderable visual evidence.

A key implication is that visual identity consistency and personality consistency should be evaluated separately. 
StoryDiffusion achieves the strongest baseline PCS, indicating strong visual stability, yet its PAS remains moderate. 
This demonstrates that mechanisms designed for appearance preservation do not automatically preserve psychological coherence. 
For narrative generation, this distinction matters: a character can remain visually recognizable while becoming behaviorally generic, emotionally overwritten, or inconsistent with their intended traits. 
Moreover, TBV's improvement in PAS is accompanied by moderate PCS reductions in several systems, suggesting that personality expressiveness and stochastic consistency are not fully orthogonal objectives. 
Future architectures should aim to resolve this tension, generating images that are both personality-aligned and generation-stable rather than trading one for the other. 
Future story-level T2I benchmarks should therefore assess not only whether a character looks the same, but also whether they continue to behave like the same psychological individual.

Our results also reveal a model-capacity pattern. 
TBV produces the largest gains in systems with stronger language-conditioning pathways, such as Qwen-Image-2512 and GPT Image 2, while CLIP-based or visual-consistency-focused systems show smaller improvements. 
This suggests that TBV does not merely add more prompt tokens; it provides a structured semantic interface that stronger text encoders can exploit more effectively. 
At the same time, the limited gains on some open-source models indicate that prompt-level grounding has an upper bound: if the text-conditioning pathway cannot preserve fine-grained behavioral cues, explicit visual constraints may still be weakened during generation.

The Emotion Override Effect further clarifies why personality consistency is difficult. 
Emotionally salient scenes significantly reduce personality alignment, indicating that transient affective context can dominate stable trait expression. 
This interaction is important because narrative characters rarely appear in neutral scenes. 
A robust story-generation model should express personality through emotion rather than replacing personality with emotion: a cautious character and an exploratory character may both feel fear, but they should not display fear in the same visual way.

Overall, TBV suggests that abstract semantic attributes must often be decomposed into observable cues before they can be rendered reliably. 
We hope future T2I work moves beyond appearance-level consistency and evaluates characters as psychologically stable agents across narrative contexts.

\clearpage 

\section*{Limitations}
\label{sec:limitations}

This work has several limitations. 
First, our evaluation relies on GPT-4o as a VLM judge for personality-related metrics. 
Although GPT-4o shows moderate agreement with human annotations and our Visual Cues Lexicon was validated by psychology-trained experts, larger-scale human evaluation is still needed across more visual styles, character types, and affective contexts.

Second, our framework is based on the Big Five (OCEAN) model, which is widely used but culturally situated. 
The visual expression of personality may vary across cultures, and some cues in the TBV Lexicon may not fully capture high-context or non-Western behavioral norms.

Third, our current evaluation focuses on one focal character per shot. 
Although the source stories may contain multiple characters, relational personality dynamics, such as dominance, avoidance, or interpersonal adaptation between characters, are not explicitly modeled.

Fourth, TBV depends on the model's ability to interpret and preserve fine-grained visual cues in the prompt. 
Its gains may therefore be limited for smaller T2I models or models with weaker text encoders, where explicit behavioral constraints can still be weakened during generation.

Finally, our temporal metrics require multiple generations per prompt and per story shot. 
This makes the evaluation computationally expensive and may limit direct extension to long-form video generation models.

\section*{Code and Data Availability}
\label{sec:availability}
The benchmark—including metadata, the Visual Cues Lexicon, prompt
templates, evaluation scripts, and anonymized annotation data—is publicly
available at \url{https://github.com/Boiyaaa/ViStoryBench-Personality}.

\section*{Ethical Considerations}
\label{sec:ethics}

This work studies how personality traits can be represented through visual behavior in generated images. 
Such capability has useful applications in storytelling, education, game design, and character-centered creative tools, but it also raises potential risks. 
More psychologically coherent generated characters may increase anthropomorphism or be misused in deceptive social interaction.

There is also a risk that mapping personality traits to visual cues could reinforce stereotypes. 
To reduce this risk, our Visual Cues Lexicon focuses on neutral and observable behavioral signals, such as posture, gaze, gesture, spatial positioning, and scene interaction. 
We avoid demographic-associated physical features and do not use protected attributes as personality indicators. 
All expert raters and human annotators participated voluntarily with informed consent. 
No monetary compensation was provided. 
No personally identifiable information was collected, and all annotations were used only in aggregate form for research evaluation.

We separately disclose the use of AI assistance in Appendix~\ref{app:ai_assistance}.

\bibliography{references}

\appendix
\section{AI Assistance Disclosure}
\label{app:ai_assistance}

AI tools were used as auxiliary support during the preparation of this paper. 
Their use was limited to literature exploration, grammar and language checking, LaTeX formatting assistance, figure/table polishing suggestions, and code-related troubleshooting.

All research questions, benchmark design, experimental settings, data construction, metric definitions, result interpretation, and scientific claims were determined and verified by the authors. 
The authors reviewed, edited, and approved the final manuscript, and take full responsibility for its content.

No AI tool was used to generate experimental results, fabricate data, alter evaluation outcomes, or replace human expert judgments. 
All reported quantitative results are derived from the described experimental pipeline and validation procedures.

\section{VLM Judge Prompts}
\label{app:vlm-prompts}

This appendix lists the prompts and decoding parameters used for VLM-based evaluation. All prompts are sent as user messages together with the evaluated image; no separate system-role message is used. The API is called with \texttt{response\_format=\{"type": "json\_object"\}} to enforce JSON-formatted outputs.

\lstset{
  basicstyle=\scriptsize\ttfamily,
  breaklines=true,
  breakatwhitespace=true,
  frame=single,
  xleftmargin=2pt,
  xrightmargin=2pt,
  aboveskip=6pt,
  belowskip=6pt
}

\subsection{PAS Task: OCEAN Annotation}

The PAS score is computed as the centered cosine similarity between the VLM-inferred OCEAN vector and the preset OCEAN vector.

\begin{lstlisting}
You are a personality psychology expert analyzing a storybook illustration.
Scene: {scene_text}
Plot: {plot_text}
Focus on these character cues (even for non-human characters like animals):
- Interaction style: approaching vs retreating, embracing vs distant
- Spatial positioning: center/dominant vs peripheral/small
- Action energy: dynamic/expansive vs still/contained
- Emotional expression: open/warm vs closed/tense (in posture and gesture, not just face)
- Overall character demeanor conveyed by the illustration
Rate each Big Five dimension from 1 (very low) to 7 (very high).
Do NOT default to 4 -- commit to a direction based on what you observe.
Respond ONLY with JSON:
{"observations": "...", "O": 4, "C": 4, "E": 4, "A": 4, "N": 4, "reasoning": "..."}
\end{lstlisting}

\subsection{RS Task: Blind Personality Inference}

The RS task evaluates whether personality can be inferred from the generated image without prior personality labels.

\begin{lstlisting}
You are a personality psychology expert. Look at this storybook illustration.
WITHOUT any prior personality information, analyze ONLY the visual cues:
Plot context (NO personality labels): {plot_text}
Focus on:
- How the character interacts with others or the environment
- Body posture: open/closed, dynamic/static, approaching/retreating
- Spatial dominance: center vs edge, large vs small
- Emotional tone conveyed through gesture and composition
Infer Big Five traits from 1 (very low) to 7 (very high).
Do NOT default to 4 -- commit to a direction.
Respond ONLY with JSON:
{"visual_evidence": "...", "inferred_O": 4, "inferred_C": 4, "inferred_E": 4, "inferred_A": 4, "inferred_N": 4, "confidence": "medium", "reasoning": "..."}
\end{lstlisting}

\subsection{SCS-Reasoning Task}

The SCS-Reasoning prompt evaluates whether the generated image reflects the expected personality-driven behavior chain.

\begin{lstlisting}
You are evaluating whether a generated image reflects a character's personality-driven behavior chain.
Plot: {plot_text}
Scene: {scene_text}
Reasoning trace (personality -> behavior -> expected visual):
{reasoning_trace}
Check:
1. Does the character's appearance match the predicted behavioral outcome?
2. Are specific visual manifestations from the trace present?
3. Rate alignment 0.0-1.0
Respond ONLY with JSON:
{"elements_found": ["..."], "elements_missing": ["..."], "alignment_score": 0.0, "rationale": "..."}
\end{lstlisting}

\subsection{VLM Generation Parameters}

All evaluations use \texttt{gpt-4o} with temperature $T{=}0.2$, \texttt{max\_completion\_tokens}$\,{=}\,$2000, $n{=}1$, high image detail, JSON response format, streaming disabled, 120\,s timeout, and up to 3 retries.

\subsection{Phase T: Trait Anchoring Prompt}
\label{app:tbv_phase_t_prompt}

Used in Phase T for each of the $k{=}5$ independent OCEAN inferences per character.

\begin{lstlisting}
You are a personality psychology expert. Given the following character
description and narrative context, infer the character's Big Five (OCEAN)
personality profile.
Character: {character_description}
Narrative context: {story_synopsis}
Rate each dimension from 1 (very low) to 7 (very high).
Do NOT default to 4 -- commit to a direction based on the narrative evidence.
Respond ONLY with JSON:
{"O": 4, "C": 4, "E": 4, "A": 4, "N": 4, "reasoning": "..."}
\end{lstlisting}

When NDI $\in [0.15, 0.33]$, the following two-round debate critic prompt is appended:

\begin{lstlisting}
The following OCEAN estimate was produced for this character: {estimate}.
Identify any dimensions where the estimate seems inconsistent with the
narrative evidence and propose corrections with justification.
Respond ONLY with JSON:
{"revised_O": 4, ..., "revised_N": 4, "critique": "..."}
\end{lstlisting}

\subsection{Phase B: Behavioral Deduction Prompt}
\label{app:tbv_phase_b_prompt}

Used in Phase B to generate the behavioral reasoning trace $r$.

\begin{lstlisting}
You are a character behavior analyst. Given a character's personality profile
and a narrative scene, describe specifically how the character would behave
in this scene. Your description must reflect the character's stable
personality, not just the emotional tone of the scene.
OCEAN profile: {ocean_vector}
Scene: {scene_description}
Respond ONLY with JSON:
{"behavioral_trace": "...", "key_behaviors": ["...", "..."],
 "reasoning": "..."}
\end{lstlisting}

The critic prompt for reverse inference (behavioral deviation check):

\begin{lstlisting}
Given this behavioral description: {trace}, infer the implied OCEAN profile
of the character exhibiting this behavior.
Respond ONLY with JSON:
{"O": 4, "C": 4, "E": 4, "A": 4, "N": 4, "reasoning": "..."}
\end{lstlisting}

\subsection{Phase V: Vision Mapping Prompt}
\label{app:tbv_phase_v_prompt}

Used in Phase V to select and compose the final visual constraint string.

\begin{lstlisting}
Given the following behavioral description and the character's OCEAN profile,
select the most relevant visual cues from the TBV Lexicon and compose them
into a visual constraint string suitable for injection into a T2I prompt.
Behavioral trace: {behavioral_trace}
OCEAN profile: {ocean_vector}
Available cues (tier-filtered): {lexicon_subset}
Apply the priority rule: facial/gaze cues > posture cues >
clothing/scene cues. Return 4-6 non-conflicting cues.
Respond ONLY with JSON:
{"selected_cues": ["...", "..."], "constraint_string": "..."}
\end{lstlisting}

\section{Independent-Judge Robustness}
\label{app:judge_robustness}

To assess whether our findings are specific to GPT-4o, we re-evaluated
a stratified subset with Gemini~2.5~Flash as an alternative judge. For each of SDXL, FLUX.1-dev, and Qwen-Image-2512, we sampled 80 matched
story shots that were generated under both C1 and C4, yielding 80 paired
observations per model. The sample was
stratified by macro-emotion category to match the affective distribution
of the full evaluation set. Every image was re-scored for PAS and RS
using the same judge prompts as the primary evaluation
(Appendix~\ref{app:vlm-prompts}); only the VLM judge was changed from
GPT-4o to Gemini~2.5~Flash.

Table~\ref{tab:judge_robustness_full} reports paired PAS gains under
both judges on the same shots. Gains agree in direction on all three
models; the two significant gains under Gemini (Qwen-Image-2512,
FLUX.1-dev) coincide with the models for which GPT-4o also reports the
largest effects. Per-shot PAS scores from the two judges correlate at
Pearson $r=.601$, close to the VLM--human agreement of $\rho=.611$ reported in
Appendix~\ref{app:vlm-human}. This suggests that the moderate absolute
agreement is characteristic of single-image personality inference in
general, rather than a limitation specific to any one judge.

This robustness analysis is a confirmatory test of the direction already
established by our primary evaluation (Section~\ref{subsec:main_results}):
that TBV improves PAS over C1. We therefore use a one-sided Wilcoxon
test, with the direction fixed before inspecting the Gemini scores.

\begin{table}[t]
\centering
\small
\setlength{\tabcolsep}{4pt}
\begin{tabular}{lcccc}
\toprule
\textbf{Model} & \textbf{Gemini $\Delta$} & \textbf{$p$} & \textbf{GPT-4o $\Delta$} & \textbf{$n$} \\
\midrule
Qwen-Image-2512 & $+0.073$ & $.021$ & $+0.075$ & 80 \\
FLUX.1-dev      & $+0.059$ & $.040$ & $+0.091$ & 80 \\
SDXL            & $+0.065$ & .133   & $+0.063$ & 80 \\
\bottomrule
\end{tabular}
\caption{PAS gains (C4$-$C1) on the same sampled shots under the
independent judge (Gemini~2.5~Flash) and the primary judge (GPT-4o).
$p$ from paired one-sided Wilcoxon signed-rank tests on the Gemini
scores. Directions agree across all models.}
\label{tab:judge_robustness_full}
\end{table}

\section{Expert Validation of the Visual Cues Lexicon}
\label{app:expert-validation}

We recruited 18 raters with graduate-level training in psychology, including Master's and doctoral students specializing in personality or clinical psychology with demonstrated familiarity with the Big Five model and affective theories. 
Raters were recruited through academic research networks and psychology graduate student mailing lists. 
Experts completed three evaluation tasks independently without discussion.

Before participation, all raters were informed that their responses would be used for research purposes to validate the Visual Cues Lexicon and the TBV pipeline. 
The instructions explained that their judgments would be analyzed only in aggregate form and reported anonymously. 
Participation was voluntary, and no personally identifiable information was collected. 
By completing the validation tasks, raters provided informed consent for their anonymized responses to be used in this study.

\subsection{Task 1: Dimension Mapping}

Each expert was presented with a visual cue description (e.g., ``character leans forward with open arms'') and asked to classify it into one of 10 OCEAN poles (O+, O$-$, C+, C$-$, E+, E$-$, A+, A$-$, N+, N$-$). The task format is illustrated below:

\begin{tcolorbox}[colback=gray!5, colframe=gray!50, boxrule=0.3pt, left=4pt, right=4pt, top=4pt, bottom=4pt, title={\small\textbf{Task 1 Example Item}}]
{\small
\textbf{Visual cue:} ``The character stands with arms crossed, avoids eye contact, and maintains a large interpersonal distance.''

\textbf{Question:} Which OCEAN pole does this cue most strongly indicate?

\textbf{Options:} O+\enspace O$-$\enspace C+\enspace C$-$\enspace E+\enspace E$-$\enspace A+\enspace A$-$\enspace N+\enspace N$-$
}
\end{tcolorbox}

\noindent Aggregated results ($n = 18$):

\begin{itemize}[nosep]
    \item Exact-match hit rate: 56.7\% (chance $= 10\%$, lift $= 5.7\times$)
    \item Dimension-level accuracy (ignoring polarity): 58.3\% (chance $= 20\%$, lift $= 2.9\times$)
    \item Fleiss' $\kappa$ (10 categories): 0.757 (substantial agreement)
\end{itemize}

\noindent Table~\ref{tab:per-pole} reports per-pole accuracy.

\begin{table}[h]
\centering
\small
\begin{tabular}{@{}lccc@{}}
\toprule
\textbf{Pole} & \textbf{Hit Rate (\%)} & \textbf{\# Cues} & \textbf{Lift} \\
\midrule
O+ & 68.8 & 16 & 6.9$\times$ \\
O$-$ & 20.3 & 15 & 2.0$\times$ \\
C+ & 78.3 & 15 & 7.8$\times$ \\
C$-$ & 74.2 & 16 & 7.4$\times$ \\
E+ & 67.2 & 16 & 6.7$\times$ \\
E$-$ & 68.8 & 14 & 6.9$\times$ \\
A+ & 67.0 & 15 & 6.7$\times$ \\
A$-$ & 34.4 & 14 & 3.4$\times$ \\
N+ & 51.0 & 14 & 5.1$\times$ \\
N$-$ & 33.7 & 15 & 3.4$\times$ \\
\bottomrule
\end{tabular}
\caption{Per-pole exact-match accuracy ($n = 18$). All poles exceed the 10\% chance baseline. Low poles of O, A, and N show lower accuracy, reflecting genuine psychological overlap between conceptually adjacent dimensions (e.g., O$-$ frequently mapped to C+).}
\label{tab:per-pole}
\end{table}

\noindent Table~\ref{tab:signal-type} reports accuracy grouped by visual signal type.

\begin{table}[h]
\centering
\small
\begin{tabular}{@{}lcc@{}}
\toprule
\textbf{Signal Type} & \textbf{Hit Rate (\%)} & \textbf{\# Cues} \\
\midrule
Facial expression & 48.4 & 26 \\
Body posture & 59.0 & 25 \\
Clothing & 63.2 & 34 \\
Color palette & 61.4 & 15 \\
Scene environment & 54.5 & 50 \\
\bottomrule
\end{tabular}
\caption{Dimension mapping accuracy by signal type. Clothing cues show the highest discriminability; facial expression cues are most ambiguous.}
\label{tab:signal-type}
\end{table}

\subsection{Task 2: Trait-Behavior Concordance}

Each expert rated the semantic fit between a personality trait and its corresponding behavioral description on a 5-point Likert scale. The task format:

\begin{tcolorbox}[colback=gray!5, colframe=gray!50, boxrule=0.3pt, left=4pt, right=4pt, top=4pt, bottom=4pt, title={\small\textbf{Task 2 Example Item}}]
{\small
\textbf{Trait:} High Agreeableness (A+)

\textbf{Behavioral description:} ``The character approaches others with open posture, maintains warm eye contact, and positions themselves at close interpersonal distance.''

\textbf{Question:} How well does this behavioral description reflect the given personality trait?

\textbf{Scale:} 1 (Strongly Disagree) -- 2 (Disagree) -- 3 (Neutral) -- 4 (Agree) -- 5 (Strongly Agree)
}
\end{tcolorbox}
\noindent Aggregated results ($n = 18$, 50 items):
\begin{itemize}[nosep]
    \item Mean rating: 4.85 ($SD = 0.40$)
    \item Percentage rated $\geq 4$ (Agree or Strongly Agree): 98.3\%
    \item S-CVI/Ave: 0.978 (excellent, threshold $\geq 0.80$)
    \item Items with I-CVI $\geq 0.78$: 50/50 (100\%)
\end{itemize}

\subsection{Task 3: Scenario Coherence}

Experts rated 30 representative scenarios---six per Russell's macro-emotion quadrant (High-Arousal-Positive, High-Arousal-Negative, Low-Arousal-Positive, Low-Arousal-Negative) plus six Neutral---for end-to-end pipeline coherence. Each scenario presented a character's OCEAN profile, a narrative scene, and the complete TBV-generated visual cue fragment. Experts rated whether the visual cues were psychologically plausible for the given character and context. The task format:

\begin{tcolorbox}[colback=gray!5, colframe=gray!50, boxrule=0.3pt, left=4pt, right=4pt, top=4pt, bottom=4pt, title={\small\textbf{Task 3 Example Item}}]
{\small
\textbf{Character profile:} O=5, C=3, E=6, A=5, N=2 (high extraversion, low neuroticism)

\textbf{Scene (High-Arousal-Positive):} ``The character wins an unexpected award at a community gathering.''

\textbf{TBV-generated visual cues:} ``The character steps forward with wide-open arms, beaming smile, upright and expansive posture, direct eye contact with the audience, animated hand gestures, relaxed facial muscles with no tension.''

\textbf{Question:} How well do these visual cues reflect a psychologically plausible response for this character in this scene?

\textbf{Scale:} 1 (Strongly Disagree) -- 2 (Disagree) -- 3 (Neutral) -- 4 (Agree) -- 5 (Strongly Agree)
}
\end{tcolorbox}

\noindent Aggregated results ($n = 18$, 30 scenarios):

\begin{itemize}[nosep]
    \item Mean coherence rating: 4.80 ($SD = 0.46$)
    \item Percentage rated $\geq 4$: 97.5\%
    \item S-CVI/Ave: 0.975 (excellent)
\end{itemize}

\noindent All thirty scenarios received unanimous or near-unanimous endorsement, confirming that the TBV pipeline maintains psychological plausibility across diverse emotional contexts.

\subsection{Effect Sizes}

To contextualize the magnitude of expert agreement:

\begin{itemize}[nosep]
    \item \textbf{Task 1 vs.\ chance:} Cohen's $h = 1.068$ (large effect), confirming that expert classification performance far exceeds random guessing.
    \item \textbf{Task 2 vs.\ neutral midpoint (3.0):} One-sample $t$-test: $t(899) = 138.9$, $p < .001$, Cohen's $d = 4.63$ (large effect), indicating concordance ratings are strongly skewed toward agreement.
\end{itemize}

\section{SCS Weight Sensitivity}
\label{app:scs_sensitivity}

We set $\alpha=0.4$ in the SCS formula to emphasize context-specific behavioral coherence over global trait recovery, reflecting the principle that story-level personality evaluation should prioritize \emph{how} a character behaves in a given scene over whether the overall trait profile is recovered.

To verify that this choice does not affect our conclusions, we swept $\alpha \in \{0.2, 0.3, 0.4, 0.5, 0.6, 0.7, 0.8\}$ and recomputed model rankings. As shown in Table~\ref{tab:scs_sensitivity}, model rankings remain identical for $\alpha \in [0.3, 0.7]$ (Kendall's $\tau = 1.0$, $p < .001$). At the extremes ($\alpha = 0.2$ and $0.8$), $\tau = 0.905$ ($p < .01$), with only the two lowest-ranked models (SDXL and StoryDiffusion) exchanging positions. These results confirm that our conclusions are not sensitive to the weighting choice.

\begin{table}[h]
\centering
\small
\begin{tabular}{ccc}
\toprule
$\alpha$ & Kendall's $\tau$ & $p$ \\
\midrule
0.2 & 0.905 & .003 \\
0.3 & 1.000 & $<$.001 \\
0.4 & 1.000 & --- \\
0.5 & 1.000 & $<$.001 \\
0.6 & 1.000 & $<$.001 \\
0.7 & 1.000 & $<$.001 \\
0.8 & 0.905 & .003 \\
\bottomrule
\end{tabular}
\caption{Kendall's $\tau$ between model rankings at each $\alpha$ and the paper's $\alpha=0.4$. Rankings are fully stable across the central range.}
\label{tab:scs_sensitivity}
\end{table}

\section{Full Visual Cues Lexicon}
\label{app:lexicon}

This appendix presents the complete set of 150 visual cues used in our annotation framework, organized by OCEAN personality dimensions and polarity. Tier annotations are C = Core ($\geq$5 / $\leq$3), E = Extended ($\geq$6 / $\leq$2), and F = Fine ($=$7 / $=$1). Cue types are Fa = facial expression, Bo = body posture, Cl = clothing, Co = color palette, and Sc = scene environment.

\subsection{Openness to Experience}
\subsubsection{High Openness}
\begingroup
\footnotesize
\renewcommand{\arraystretch}{1}
\begin{tabular}{@{}c c p{5.2cm}@{}}
\toprule
\textbf{Tier} & \textbf{Type} & \textbf{Visual cue} \\
\midrule
C & Fa & Animated or expressive facial micro-movements \\
C & Fa & Wide open or exploratory gaze \\
C & Bo & Forward-leaning curious posture \\
C & Bo & Relaxed open hand gestures \\
C & Bo & Slight head tilt suggesting interest \\
E & Cl & Layered or mixed-texture clothing \\
E & Cl & Asymmetric or flowing silhouettes \\
E & Cl & Artistic or handcrafted accessories \\
E & Co & Warm or vivid accent colors (mustard, rust, teal) \\
E & Cl & Unconventional or eclectic outfit combinations \\
F & Sc & Bookshelf, map, or artwork in background \\
F & Sc & Natural outdoor light or dappled sunlight \\
F & Sc & Creative or slightly cluttered studio environment \\
F & Sc & Plants, cultural objects, or travel items as props \\
F & Sc & Warm diffuse ambient lighting \\
\bottomrule
\end{tabular}
\endgroup

\subsubsection{Low Openness}
\begingroup
\footnotesize
\renewcommand{\arraystretch}{1}
\begin{tabular}{@{}c c p{5.2cm}@{}}
\toprule
\textbf{Tier} & \textbf{Type} & \textbf{Visual cue} \\
\midrule
C & Fa & Composed neutral or flat facial expression \\
C & Fa & Forward-facing steady gaze, no exploratory movement \\
C & Bo & Upright symmetrical stance \\
C & Bo & Arms close to body or hands clasped \\
C & Bo & Deliberate controlled movements, no fidgeting \\
E & Cl & Well-fitted traditional or classic silhouette \\
E & Cl & No accessories or minimal functional accessories \\
E & Cl & Conventional or uniform-style clothing \\
E & Co & Muted neutral palette (navy, grey, beige, black) \\
E & Co & Single-color or subtle pattern outfit \\
F & Sc & Plain or uniform background wall \\
F & Sc & Orderly indoor space with consistent lighting \\
F & Sc & Minimal props, no clutter \\
F & Sc & Flat even artificial lighting \\
F & Sc & No decorative or cultural objects visible \\
\bottomrule
\end{tabular}
\endgroup

\subsection{Conscientiousness}

\subsubsection{High Conscientiousness}
\begingroup
\footnotesize
\renewcommand{\arraystretch}{1}
\begin{tabular}{@{}c c p{5.2cm}@{}}
\toprule
\textbf{Tier} & \textbf{Type} & \textbf{Visual cue} \\
\midrule
C & Fa & Subtle confident closed-lip expression, no tension \\
C & Fa & Focused and direct gaze \\
C & Bo & Tall upright posture with engaged core \\
C & Bo & Deliberate purposeful hand placement, no fidgeting \\
C & Bo & Still body with minimal movement \\
E & Cl & Neatly pressed formal or semi-formal attire \\
E & Cl & Classic or professional silhouette \\
E & Cl & Minimal but purposeful accessories \\
E & Co & Coordinated color palette, no clashing tones \\
E & Cl & Well-maintained fabric with no wrinkles \\
F & Sc & Clean organized desk or workspace in background \\
F & Sc & Planner, checklist, or clock visible as prop \\
F & Sc & Neat filing or structured shelving in background \\
F & Sc & Consistent balanced overhead lighting \\
F & Sc & Nothing out of place in background \\
\bottomrule
\end{tabular}
\endgroup

\subsubsection{Low Conscientiousness}
\begingroup
\footnotesize
\renewcommand{\arraystretch}{1}
\begin{tabular}{@{}c c p{5.2cm}@{}}
\toprule
\textbf{Tier} & \textbf{Type} & \textbf{Visual cue} \\
\midrule
C & Fa & Relaxed jaw, slightly open mouth, unfocused expression \\
C & Fa & Wandering or unfocused gaze \\
C & Bo & Slouched or leaning posture \\
C & Bo & Hands fidgeting or in pockets \\
C & Bo & Weight shifted to one side, asymmetric stance \\
E & Cl & Wrinkled or untucked casual clothing \\
E & Cl & Shirt untucked or collar askew \\
E & Cl & Comfort-first athleisure or oversized fit \\
E & Co & Mismatched or random color combinations \\
E & Cl & Worn or slightly disheveled items \\
F & Sc & Cluttered or disorganized background \\
F & Sc & Scattered objects or papers visible \\
F & Sc & Unmade surface or piled items in background \\
F & Sc & Dim or inconsistent ambient lighting \\
F & Sc & Items visibly out of place in background \\
\bottomrule
\end{tabular}
\endgroup

\subsection{Extraversion}

\subsubsection{High Extraversion}
\begingroup
\footnotesize
\renewcommand{\arraystretch}{1}
\begin{tabular}{@{}c c p{5.2cm}@{}}
\toprule
\textbf{Tier} & \textbf{Type} & \textbf{Visual cue} \\
\midrule
C & Fa & Broad confident smile showing teeth (Duchenne smile) \\
C & Fa & Direct sustained eye contact \\
C & Bo & Arms extended outward, open gesturing \\
C & Bo & Erect expansive posture occupying space \\
C & Bo & Body oriented toward viewer or others \\
E & Cl & Attention-drawing patterns or prints \\
E & Cl & Open necklines or expressive cuts \\
E & Cl & Bold statement accessories \\
E & Co & Bright warm colors (red, orange, yellow, warm pink) \\
E & Cl & Fitted clothing that emphasizes physical presence \\
F & Sc & Social gathering or crowd in background \\
F & Sc & Bright high-key lighting \\
F & Sc & Open spacious environment \\
F & Sc & Stage, podium, or center-of-room position \\
F & Sc & Lively colorful setting \\
\bottomrule
\end{tabular}
\endgroup

\subsubsection{Low Extraversion (Introversion)}
\begingroup
\footnotesize
\renewcommand{\arraystretch}{1}
\begin{tabular}{@{}c c p{5.2cm}@{}}
\toprule
\textbf{Tier} & \textbf{Type} & \textbf{Visual cue} \\
\midrule
C & Fa & Minimal facial expression, closed mouth, neutral affect \\
C & Fa & Downward or averted gaze \\
C & Bo & Arms crossed or drawn close to body \\
C & Bo & Body angled away from viewer \\
C & Bo & Weight inward, occupying minimal space \\
E & Cl & Close-fitting minimal silhouette, no statement elements \\
E & Cl & Covered or layered clothing \\
E & Cl & No statement accessories \\
E & Co & Dark muted tones (charcoal, navy, deep grey) \\
E & Cl & Understated fabric textures \\
F & Sc & Quiet solitary corner or alcove \\
F & Sc & Dim soft interior lighting \\
F & Sc & Single chair or isolated seating \\
F & Sc & Window with soft natural light, solitary setting \\
F & Sc & Few or no other people visible in background \\
\bottomrule
\end{tabular}
\endgroup

\subsection{Agreeableness}

\subsubsection{High Agreeableness}
\begingroup
\footnotesize
\renewcommand{\arraystretch}{1}
\begin{tabular}{@{}c c p{5.2cm}@{}}
\toprule
\textbf{Tier} & \textbf{Type} & \textbf{Visual cue} \\
\midrule
C & Fa & Gentle closed-lip or soft open smile \\
C & Fa & Slight head tilt indicating empathy \\
C & Bo & Open palms facing upward \\
C & Bo & Body leaning slightly forward toward others \\
C & Fa & Soft relaxed brow, no tension \\
E & Cl & Comfortable relaxed-fit clothing \\
E & Cl & Rounded gentle silhouettes, no sharp styling \\
E & Cl & Natural or organic fabric textures \\
E & Co & Soft warm tones (peach, cream, sage, dusty rose) \\
E & Cl & No sharp or aggressive styling elements \\
F & Sc & Warm cozy interior with soft furnishings \\
F & Sc & Plants, flowers, or natural elements as props \\
F & Sc & Warm golden or candlelit ambient lighting \\
F & Sc & Collaborative or shared space in background \\
F & Sc & Soft textures like cushions or throws visible \\
\bottomrule
\end{tabular}
\endgroup

\subsubsection{Low Agreeableness}
\begingroup
\footnotesize
\renewcommand{\arraystretch}{1}
\begin{tabular}{@{}c c p{5.2cm}@{}}
\toprule
\textbf{Tier} & \textbf{Type} & \textbf{Visual cue} \\
\midrule
C & Fa & Sharp sustained eye contact, unblinking \\
C & Fa & Tense jaw or clenched teeth visible \\
C & Fa & Flat or downturned expression \\
C & Bo & Arms crossed high on chest or hands on hips \\
C & Bo & Body square-on, space-assertive stance \\
E & Cl & Sharp angular silhouettes or structured cuts \\
E & Cl & Stiff formal or militaristic styling \\
E & Cl & Power-signaling or imposing accessories \\
E & Co & Cold tones (black, steel grey, icy blue) \\
E & Co & High-contrast color blocking \\
F & Sc & Stark industrial or minimal background \\
F & Sc & Hard directional lighting creating strong shadows \\
F & Sc & Empty impersonal space \\
F & Sc & No decorative elements, bare walls \\
F & Sc & Cool color temperature lighting \\
\bottomrule
\end{tabular}
\endgroup

\subsection{Neuroticism}

\subsubsection{High Neuroticism}
\begingroup
\footnotesize
\renewcommand{\arraystretch}{1}
\begin{tabular}{@{}c c p{5.2cm}@{}}
\toprule
\textbf{Tier} & \textbf{Type} & \textbf{Visual cue} \\
\midrule
C & Fa & Furrowed brow or creased forehead \\
C & Fa & Tight pressed lips or bitten lip \\
C & Fa & Downcast or darting anxious gaze \\
C & Bo & Raised and tensed shoulders \\
C & Bo & Hands wringing, fidgeting, or gripping \\
E & Cl & Disheveled or slightly unkempt presentation \\
E & Cl & Clothing that looks pulled at or self-touched \\
E & Cl & Self-concealing layered or oversized items \\
E & Co & Dark or dull desaturated colors \\
E & Co & Mismatched or incoherent outfit choices \\
F & Sc & Dark shadowy or low-key environment \\
F & Sc & Confined or enclosed space \\
F & Sc & Dramatic or harsh lighting with strong shadows \\
F & Sc & Visually chaotic or disordered background \\
F & Sc & Isolated setting with no comfort elements \\
\bottomrule
\end{tabular}
\endgroup

\subsubsection{Low Neuroticism (Emotional Stability)}
\begingroup
\footnotesize
\renewcommand{\arraystretch}{1}
\begin{tabular}{@{}c c p{5.2cm}@{}}
\toprule
\textbf{Tier} & \textbf{Type} & \textbf{Visual cue} \\
\midrule
C & Fa & Calm smooth brow with no creasing \\
C & Fa & Steady unhurried gaze \\
C & Bo & Relaxed dropped shoulders \\
C & Bo & Easy natural hand placement at sides \\
C & Bo & Slow deliberate composed movements \\
E & Cl & Neat but relaxed presentation \\
E & Cl & No signs of self-touching or garment distortion \\
E & Cl & Well-coordinated comfortable outfit \\
E & Co & Clean neutral or balanced tones (soft white, light grey, calm blue) \\
E & Cl & Natural fabric, nothing stiff or aggressive \\
F & Sc & Bright airy open environment \\
F & Sc & Natural daylight or soft warm light \\
F & Sc & Calm water, greenery, or nature elements \\
F & Sc & Clean uncluttered background \\
F & Sc & Spacious setting with comfortable proportions \\
\bottomrule
\end{tabular}
\endgroup

\section{Dataset Statistics}
\label{app:dataset_stats}

ViStoryBench-Personality is built on the narrative foundation of ViStoryBench \citep{zhuang2025vistory}, spanning 80 stories and 1,317 sequential shots. Table~\ref{tab:dataset_overview} summarizes the key statistics.

\begin{table}[h]
\centering
\small
\begin{tabular}{@{}lr@{}}
\toprule
\textbf{Statistic} & \textbf{Value} \\
\midrule
Total stories & 80 \\
Total sequential shots & 1,317 \\
Unique characters & 80 \\
Mean shots per story & 16.5 \\
Evaluation shots & 213 \\
Conditions per shot & 4 (C1--C4) \\
T2I models evaluated & 7 \\
Total generated images & 15,336 \\
\bottomrule
\end{tabular}
\caption{Overview of the ViStoryBench-Personality dataset.}
\label{tab:dataset_overview}
\end{table}

\paragraph{Evaluation Subset.}
The 213 evaluation shots were selected via balanced sampling across five affective scene categories derived from Russell's Circumplex Model \citep{russell1980circumplex}. 
This design ensures that the evaluation covers both neutral and emotionally salient contexts, allowing us to analyze whether scene-level affect overrides personality-specific visual cues. 
Table~\ref{tab:emotion_distribution} reports the affective category and visual difficulty distribution.

\begin{table}[h]
\centering
\small
\begin{tabular}{@{}lcc@{}}
\toprule
\textbf{Category} & \textbf{Count} & \textbf{\%} \\
\midrule
Neutral & 44 & 20.7 \\
Low-arousal positive & 41 & 19.2 \\
Low-arousal negative & 42 & 19.7 \\
High-arousal positive & 40 & 18.8 \\
High-arousal negative & 46 & 21.6 \\
\midrule
\multicolumn{3}{@{}l}{\textit{Visual difficulty}} \\
\quad Direct (facial/gestural) & 153 & 71.8 \\
\quad Indirect (spatial/implicit) & 60 & 28.2 \\
\bottomrule
\end{tabular}
\caption{Distribution of affective scene categories and visual difficulty across 213 evaluation shots.}
\label{tab:emotion_distribution}
\end{table}

\begin{figure}[h]
\centering
\includegraphics[width=\columnwidth]{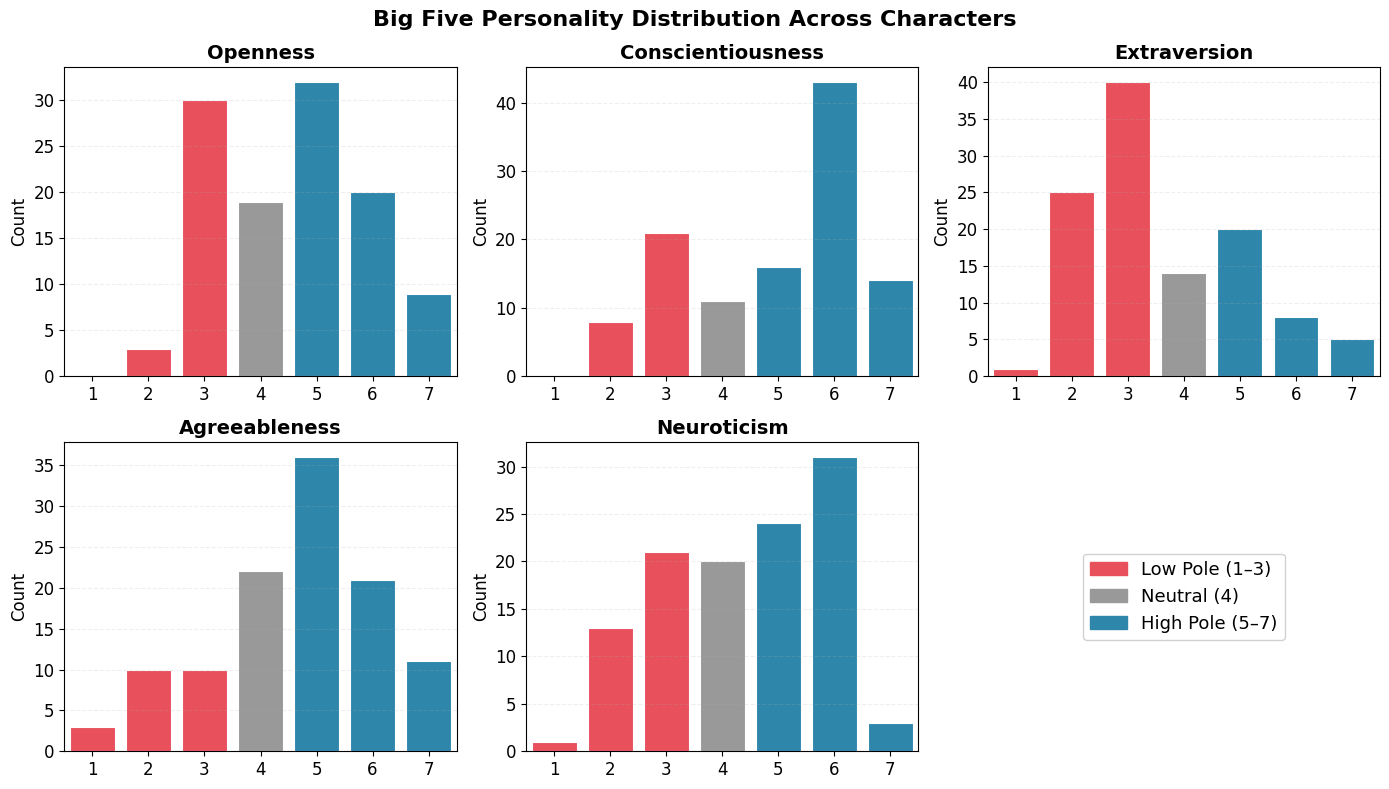}
\caption{Big Five personality distribution across unique characters. Red = low pole (1--3), grey = neutral (4), blue = high pole (5--7).}
\label{fig:ocean_distribution}
\end{figure}

\paragraph{OCEAN Distribution.}
Figure~\ref{fig:ocean_distribution} shows the Big Five personality distribution across unique characters. 
The distribution reflects the narrative archetypes in the source corpus: high Conscientiousness is overrepresented, corresponding to goal-driven protagonists in adventure stories, while extreme low-pole scores are less frequent. 
Neutral scores also appear across all dimensions, providing a natural control group for evaluating whether T2I models default to generic personality expression when no strong directional trait signal is present.

\section{Prompt Construction Example}
\label{app:prompt_example}

Table~\ref{tab:prompt_example_appendix} shows a representative prompt construction example across the four ablation conditions.

\begin{table}[h]
\centering
\footnotesize
\setlength{\tabcolsep}{3pt}
\begin{tabular}{@{}cp{5.8cm}@{}}
\toprule
\textbf{Cond.} & \textbf{Prompt content} \\
\midrule
C1 & Alice wishes she could shrink to fit through the small door. Daytime, inside the hall, disappointed and hopeful atmosphere. A curious girl with golden hair and blue eyes. \\
C2 & C1 + \textit{[Personality: High Openness, Low Conscientiousness, High Extraversion, Low Neuroticism]} \\
C3 & C2 + \textit{feeling Disappointment} \\
C4 & C1 + \textit{animated facial micro-movements $|$ wide open exploratory gaze $|$ forward-leaning curious posture $|$ relaxed open hand gestures $|$ slight head tilt suggesting interest} \\
\bottomrule
\end{tabular}
\caption{Example prompt construction across four abstraction levels for Alice (O=7, C=3, E=5, A=4, N=3). C1--C3 test non-visual textual information, while C4 tests concrete visual grounding.}
\label{tab:prompt_example_appendix}
\end{table}
\section{Implementation Details}
\label{app:implementation}

\subsection{T2I Model Configurations}

Table~\ref{tab:model_config} summarizes the generation parameters for all seven models. Open-source models were run locally; closed-source models were accessed via their respective APIs.

\begin{table}[h]
\centering
\footnotesize
\setlength{\tabcolsep}{3pt}
\begin{tabular}{@{}lccll@{}}
\toprule
\textbf{Model} & \textbf{Steps} & \textbf{CFG} & \textbf{Resolution} & \textbf{Arch.} \\
\midrule
SDXL & 28 & 7.5 & 1024$\times$1024 & UNet+CLIP \\
FLUX.1-dev & 28 & 3.5 & 1024$\times$1024 & DiT+T5 \\
SD3.5-Large & 28 & 4.5 & 1024$\times$1024 & MMDiT+3Enc \\
Qwen-2512 & 50 & 4.0 & 1328$\times$1328 & MMDiT+LLM \\
StoryDiff. & 50 & 5.0 & 1024$\times$1024 & SDXL+CSA \\
GPT Image 2 & --- & --- & 1254$\times$1254 & Closed \\
Nano Banana & --- & --- & 1408$\times$768 & Closed \\
\bottomrule
\end{tabular}
\caption{Generation parameters for the seven T2I models.}
\label{tab:model_config}
\end{table}

\subsection{Hardware and Runtime}
Open-source models were run on NVIDIA RTX 4090 (24\,GB) or A100 (80\,GB). Qwen-2512 and StoryDiffusion were run on A100 due to higher memory requirements; remaining open-source models used RTX 4090 with \texttt{sequential\_cpu\_offload}. Table~\ref{tab:runtime} reports approximate per-image generation time.

\begin{table}[h]
\centering
\small
\begin{tabular}{@{}p{2.5cm}cc@{}}
\toprule
\textbf{Model} & \textbf{Device} & \textbf{Time/image} \\
\midrule
SDXL & RTX 4090 & $\sim$8\,s \\
FLUX.1-dev & RTX 4090 & $\sim$12\,s \\
SD3.5-Large & RTX 4090 & $\sim$15\,s \\
Qwen-2512 & A100 & $\sim$25\,s \\
StoryDiffusion & A100 & $\sim$20\,s \\
GPT Image 2 & API & $\sim$15\,s \\
Nano Banana & API & $\sim$15\,s \\
\bottomrule
\end{tabular}
\caption{Approximate per-image generation time.}
\label{tab:runtime}
\end{table}

\subsection{Generation Protocol}

Each of the 213 evaluation shots was generated under 4 ablation conditions (C1--C4). For the five open-source models (SDXL, FLUX.1-dev, SD3.5-Large, Qwen-2512, StoryDiffusion), each condition was generated 3 times with different random seeds, yielding $213 \times 4 \times 3 \times 5 = 12{,}780$ images. For the two closed-source models (GPT Image 2, Nano Banana), a total of $1{,}278 \times 2 = 2{,}556$ images were generated. The complete evaluation set comprises 15,336 generated images.

\subsection{VLM Evaluation Configuration}

Personality assessment was conducted using GPT-4o via API with the following parameters: temperature $T{=}0.2$, \texttt{max\_completion\_tokens}$\,{=}\,$2000, $n{=}1$, high image detail, and JSON response format (\texttt{response\_format=\{"type": "json\_object"\}}). Each generated image was evaluated on three tasks (PAS, RS, SCS-Reasoning), and consecutive shots within each story were evaluated for PCS.

\subsection{TBV LLM Configuration}
\label{app:tbv_llm}

All LLM-based operations in TBV (Phase T, Phase B, and Phase V) are implemented using GPT-4o (\texttt{gpt-4o-2024-11-20}) with temperature $T{=}0.2$ and \texttt{max\_completion\_tokens}$\,{=}\,$1000. Phase T samples $k{=}5$ independent OCEAN inferences per character. If the resulting NDI falls in $[0.15, 0.33]$, a two-round multi-agent debate is triggered: a second GPT-4o instance critiques the initial estimate and proposes a revised anchor, after which the mean of non-outlier estimates is accepted. Phase B applies the self-correcting loop with a maximum of three regeneration attempts before falling back to the best-scoring trace. The full prompts for all three phases are listed in Appendix~\ref{app:vlm-prompts} (Sections~\ref{app:tbv_phase_t_prompt}--\ref{app:tbv_phase_v_prompt}).

\subsection{Cost Estimation}

Table~\ref{tab:cost} summarizes the estimated API and compute cost.

\begin{table}[h]
\centering
\small
\begin{tabular}{@{}p{3.2cm}r@{}}
\toprule
\textbf{Component} & \textbf{Est.\ Cost (USD)} \\
\midrule
GPT Image 2 generation & $\sim$\$240 \\
Nano Banana generation & $\sim$\$240 \\
VLM evaluation (GPT-4o) & $\sim$\$180 \\
RTX 4090 GPU hours & $\sim$200\,h \\
A100 GPU hours & $\sim$20\,h \\
\midrule
\textbf{Total API cost} & $\sim$\textbf{\$660} \\
\bottomrule
\end{tabular}
\caption{Estimated cost breakdown for the full evaluation pipeline. GPU hours reported separately as institutional compute.}
\label{tab:cost}
\end{table}

\section{VLM-Human Agreement Details}
\label{app:vlm-human}

To validate GPT-4o as an automated personality judge, we conducted a human agreement study. 
We sampled a stratified calibration subset of 100 generated images covering multiple T2I systems, all five affective scene categories, and both direct and indirect visual-expression shots. 
Six psychology-trained annotators, all Master's or doctoral students in psychology-related fields, independently rated each image on the five OCEAN dimensions using the same 7-point scale as the VLM judge. 
Final human scores were computed as the mean rating per dimension per image. 
Annotators participated voluntarily with informed consent; no personally identifiable information was collected, and all ratings were analyzed only in aggregate form. 
We then compared GPT-4o predictions with the averaged human ratings using rank correlation, absolute error, within-one-point accuracy, and polarity agreement.

\subsection{Overall Agreement}

Table~\ref{tab:vlm-human-overall} reports aggregate agreement statistics.

\begin{table}[h]
\centering
\small
\begin{tabular}{@{}p{4.0cm}r@{}}
\toprule
\textbf{Metric} & \textbf{Value} \\
\midrule
Spearman $\rho$ & 0.611 \\
Pearson $r$ & 0.613 \\
MAE (7-point scale) & 0.871 \\
Within $\pm$1 point & 81.3\% \\
Polarity agreement & 86.6\% \\
\bottomrule
\end{tabular}
\caption{VLM-human agreement on personality evaluation. Polarity agreement measures whether VLM and humans assign the same trait direction (above vs.\ below neutral); chance $= 50\%$.}
\label{tab:vlm-human-overall}
\end{table}

The moderate score-level correlation (Spearman $\rho = 0.611$), together with 86.6\% polarity agreement, supports the use of GPT-4o as a scalable proxy for directional personality evaluation. However, the remaining disagreement indicates that automated judgments should not be interpreted as a replacement for human evaluation.

\subsection{Per-Dimension Polarity Agreement}

Table~\ref{tab:polarity-dim} reports polarity agreement by OCEAN dimension. GPT-4o and human annotators agreed on trait direction in 86.6\% of cases overall, with per-dimension agreement ranging from 77.3\% (Extraversion) to 97.6\% (Openness).

\begin{table}[h]
\centering
\small
\begin{tabular}{@{}lcc@{}}
\toprule
\textbf{Dimension} & \textbf{Agreement} & \textbf{Chance} \\
\midrule
Openness (O) & 97.6\% & 50\% \\
Conscientiousness (C) & 86.8\% & 50\% \\
Extraversion (E) & 77.3\% & 50\% \\
Agreeableness (A) & 81.2\% & 50\% \\
Neuroticism (N) & 89.1\% & 50\% \\
\midrule
\textbf{Overall} & \textbf{86.6\%} & \textbf{50\%} \\
\bottomrule
\end{tabular}
\caption{Polarity agreement between VLM and human annotators by OCEAN dimension (chance $= 50\%$). Neutral ratings ($= 4$) are excluded.}
\label{tab:polarity-dim}
\end{table}

The relatively lower agreement on Extraversion may reflect the difficulty of distinguishing stable social-energy cues from transient emotional expressions in narrative illustrations, consistent with the Emotion Override Effect observed in our main analysis.

\subsection{Score-Level vs.\ Directional Agreement}

The moderate score-level agreement ($\rho = 0.611$) combined with high polarity agreement (86.6\%) indicates that GPT-4o reliably captures the \textit{direction} of personality expression while showing greater variability in exact magnitude estimates. This pattern is well-suited to our evaluation framework, which uses centered cosine similarity and is therefore more sensitive to directional agreement than to exact score magnitude.

\section{Inter-Metric Correlations}
\label{app:metric_corr}

To assess whether the six metrics provide non-redundant signals, we
compute their pairwise Pearson correlations across all per-shot records
with complete values on all six metrics ($n = 5{,}056$; Table~\ref{tab:metric_corr}). Excluding SCS, which is a
deterministic composite ($\mathrm{SCS} = 0.4\cdot\mathrm{PAS} +
0.6\cdot\mathrm{SCS_{reasoning}}$ by definition), pairwise correlations
range from .13 to .66. PAS and RS correlate moderately ($r=.66$; 43\%
shared variance), while PCS and the reasoning channel are nearly
independent of the alignment metrics ($r\le.20$). This indicates that
the suite measures distinct facets of personality expression rather than
repackaging a single inference. CCS and PDS are sequence-level metrics and are reported per model and
condition in Table~\ref{tab:main_results}.

\begin{table}[t]
\centering
\small
\setlength{\tabcolsep}{4pt}
\begin{tabular}{lccccc}
\toprule
 & PAS & PCS & RS & SCS$_r$ & SCS \\
\midrule
PAS     & 1.00 & .16 & .66 & .14 & .83 \\
PCS     & .16 & 1.00 & .13 & .14 & .20 \\
RS      & .66 & .13 & 1.00 & .17 & .59 \\
SCS$_r$ & .14 & .14 & .17 & 1.00 & .66 \\
SCS     & .83 & .20 & .59 & .66 & 1.00 \\
\bottomrule
\end{tabular}
\caption{Pearson correlations among per-shot metrics ($n=5{,}056$).
SCS$_r$ = SCS reasoning component. SCS's high correlation with PAS and
SCS$_r$ is constructional (weighted composite).}
\label{tab:metric_corr}
\end{table}

\section{Full Ablation Results}
\label{app:ablation}

Table~\ref{tab:ablation_full} reports the complete C1--C4 ablation results for the five open-source models.

\begin{table*}[t]
\centering
\small
\caption{Full ablation study results. C1 = base narrative prompt (plot + appearance); C2 = C1 + raw OCEAN trait tags; C3 = C2 + deduced scene-level emotion labels; C4 (Full TBV) = C1 + explicit visual cues from the TBV lexicon. Appending abstract psychological vocabulary (C2, C3) yields marginal or negative effects, confirming the semantic gap. Only concrete visual grounding (C4) produces consistent gains. $n = 213$ per condition for open-source models.}
\label{tab:ablation_full}
\begin{tabular}{@{}llcccccccccc@{}}
\toprule
& & \multicolumn{2}{c}{\textbf{SDXL}} & \multicolumn{2}{c}{\textbf{StoryDiffusion}} & \multicolumn{2}{c}{\textbf{FLUX.1-dev}} & \multicolumn{2}{c}{\textbf{SD3.5-Large}} & \multicolumn{2}{c}{\textbf{Qwen-2512}} \\
\cmidrule(lr){3-4} \cmidrule(lr){5-6} \cmidrule(lr){7-8} \cmidrule(lr){9-10} \cmidrule(lr){11-12}
& Cond & PAS & RS & PAS & RS & PAS & RS & PAS & RS & PAS & RS \\
\midrule
& C1 & 0.446 & 0.429 & 0.480 & 0.467 & 0.464 & 0.469 & 0.466 & 0.493 & 0.518 & 0.546 \\
& C2 & 0.392 & 0.382 & 0.462 & 0.449 & 0.483 & 0.475 & 0.452 & 0.469 & 0.545 & 0.560 \\
& C3 & 0.408 & 0.400 & 0.471 & 0.444 & 0.484 & 0.460 & 0.460 & 0.471 & 0.540 & 0.575 \\
& C4 & 0.467 & 0.423 & 0.502 & 0.508 & 0.522 & 0.552 & 0.472 & 0.488 & 0.633 & 0.641 \\
\midrule
\multicolumn{2}{@{}l}{$\Delta$(C4--C1)} & +4.8\% & --1.4\% & +4.6\% & +8.8\% & +12.6\% & +17.7\% & +1.5\% & --1.0\% & +22.1\% & +17.4\% \\
\bottomrule
\end{tabular}
\end{table*}

\section{Per-Dimension Gain Decomposition}
\label{app:trait_breakdown}

Some lexicon poles are less reliably mapped by experts than others
(Appendix~\ref{app:expert-validation}), raising the question of whether
TBV's benefit also holds for these weaker poles. To examine this, we
decompose the C4$-$C1 improvement by OCEAN dimension and polarity. For
each dimension $d$, we use its contribution to the centered-cosine
numerator, $(v_d - m)(\hat{v}_d - m)$, as a per-dimension alignment
diagnostic aggregated across all evaluated shots. Because this quantity
is a product of centered scores rather than a cosine similarity, it is
not on the PAS scale and is reported only as a relative decomposition of
where gains arise.

All ten trait poles show positive C4$-$C1 gains. Critically, the
weakest-mapped poles in the expert study, low Openness (expert hit rate
20.3\%) and low Neuroticism (33.7\%), show gains comparable to or larger
than the most reliably mapped pole (high Conscientiousness, 78.3\%).
TBV's benefit therefore does not depend on the most discriminable cues:
even weakly mapped dimensions gain from explicit visual grounding. This
is consistent with the view that concrete rendering, rather than lexicon
reliability alone, drives the improvement.

\section{Statistical Tests for Ablation Conditions}
\label{app:stat_tests}

To test whether abstract personality descriptors and additional scene-level emotion labels produce consistent PAS improvements over the baseline prompt, we conducted paired Wilcoxon signed-rank tests between C1 and each of C2, C3, and C4 across all models ($n = 213$ per pair for open-source models). We additionally applied two one-sided tests (TOST; $\delta = 0.05$) to the C1--C2 and C1--C3 comparisons to assess practical equivalence. Table~\ref{tab:stat_tests} reports the full results.

Of the ten C1\,vs.\,C2/C3 comparisons across open-source models, nine show no significant improvement (Wilcoxon $p > .05$). The single exception is Qwen-Image-2512 under C2 ($p = .027$, $r = 0.21$), where the gain ($\Delta = +0.027$) is small in magnitude and inconsistent across models---SDXL shows a decrease of comparable magnitude under C2 ($\Delta = -0.054$, $p = .054$). TOST equivalence is confirmed for 6 of 10 comparisons; the remaining four involve either a borderline positive shift (Qwen-Image-2512 C2/C3, TOST $p = .051/.054$) or a negative shift exceeding the equivalence margin (SDXL C2/C3). This pattern indicates that abstract descriptors produce \emph{inconsistent} rather than \emph{uniformly absent} effects---neither constitutes reliable trait grounding.

In contrast, C1\,vs.\,C4 yields significant improvements for FLUX.1-dev ($p = .005$, $r = 0.24$), Qwen-Image-2512 ($p < .001$, $r = 0.43$), and both closed-source models ($p < .001$). SDXL and SD3.5-Large show non-significant C4 gains ($p = .160$ and $.634$), consistent with the text-encoder capacity pattern discussed in Section~\ref{subsec:main_results}.

\begin{table*}[t]
\centering
\small
\begin{tabular}{llccccc}
\toprule
Model & Comparison & $\Delta$PAS & Wilcoxon $p$ & $r$ & TOST $p$ & Equiv. \\
\midrule
FLUX.1-dev       & C1 vs C2 & $+$0.015 & .161 & 0.13 & .009 & \checkmark \\
                 & C1 vs C3 & $+$0.020 & .372 & 0.08 & .045 & \checkmark \\
                 & C1 vs C4 & $+$0.059 & .005 & 0.24 & --- & --- \\
\midrule
SDXL             & C1 vs C2 & $-$0.054 & .054 & $-$0.17 & .561 & $\times$ \\
                 & C1 vs C3 & $-$0.038 & .134 & $-$0.13 & .279 & $\times$ \\
                 & C1 vs C4 & $+$0.021 & .160 & 0.12 & --- & --- \\
\midrule
SD3.5-Large      & C1 vs C2 & $-$0.014 & .703 & $-$0.03 & .018 & \checkmark \\
                 & C1 vs C3 & $-$0.005 & .996 & 0.00 & .002 & \checkmark \\
                 & C1 vs C4 & $+$0.007 & .634 & 0.04 & --- & --- \\
\midrule
StoryDiffusion   & C1 vs C2 & $-$0.018 & .436 & $-$0.07 & .043 & \checkmark \\
                 & C1 vs C3 & $-$0.009 & .478 & $-$0.06 & .012 & \checkmark \\
                 & C1 vs C4 & $+$0.022 & .148 & 0.12 & --- & --- \\
\midrule
Qwen-Image-2512  & C1 vs C2 & $+$0.027 & .027$^{*}$ & 0.21 & .051 & $\times$ \\
                 & C1 vs C3 & $+$0.021 & .307 & 0.10 & .054 & $\times$ \\
                 & C1 vs C4 & $+$0.115 & ${<}.001$ & 0.43 & --- & --- \\
\midrule
GPT Image 2      & C1 vs C4 & $+$0.108 & ${<}.001$ & 0.50 & --- & --- \\
Nano Banana      & C1 vs C4 & $+$0.094 & ${<}.001$ & 0.39 & --- & --- \\
\bottomrule
\end{tabular}
\caption{Paired Wilcoxon signed-rank tests and TOST equivalence tests ($\delta = 0.05$) for PAS across ablation conditions. $r$ = rank-biserial correlation. Equiv. = TOST equivalence at $\delta = 0.05$: \checkmark{} confirmed, $\times$ not confirmed, --- not applicable. TOST is reported only for C1 vs. C2/C3 comparisons, which are the focus of the equivalence claim; C1 vs. C4 is tested as the intervention condition. $^{*}$Significant at $p<.05$. Closed-source models are evaluated under C1 and C4 only.}
\label{tab:stat_tests}
\end{table*}

\section{Emotion Override Statistical Analysis}
\label{app:emotion_override}

We analyze PAS across five affective scene categories: Neutral, Low-Arousal Positive, Low-Arousal Negative, High-Arousal Positive, and High-Arousal Negative. 
After removing records with missing PAS or affective labels, the analysis contains 5,081 valid observations. 
Table~\ref{tab:emotion_pas_summary} reports descriptive statistics for each category.

\begin{table}[h]
\centering
\small
\begin{tabular}{@{}lrrrr@{}}
\toprule
\textbf{Category} & \textbf{N} & \textbf{Mean} & \textbf{SD} & \textbf{Median} \\
\midrule
Neutral & 1053 & 0.575 & 0.311 & 0.655 \\
High-Arousal Positive & 952 & 0.507 & 0.337 & 0.571 \\
Low-Arousal Negative & 998 & 0.498 & 0.338 & 0.541 \\
High-Arousal Negative & 1097 & 0.477 & 0.374 & 0.567 \\
Low-Arousal Positive & 981 & 0.446 & 0.351 & 0.474 \\
\bottomrule
\end{tabular}
\caption{PAS distribution across affective scene categories. Neutral scenes obtain the highest mean PAS, while all emotionally salient categories show lower mean PAS.}
\label{tab:emotion_pas_summary}
\end{table}

A one-way ANOVA shows a significant effect of affective category on PAS, $F(4,5076)=20.06$, $p<.001$. 
Because Levene's test indicates unequal variances across groups ($p<.001$), we also conduct a non-parametric Kruskal--Wallis test, which confirms the same pattern ($H=82.00$, $p<.001$). 
Tukey HSD post-hoc comparisons show that Neutral differs significantly from all four emotionally salient categories, while several comparisons among emotional categories are not significant. 
This supports the Emotion Override Effect: emotionally salient contexts tend to reduce visually readable personality alignment relative to neutral contexts.

\section{Indirect Expression Gap (Null Result)}
\label{app:indirect}

We hypothesized that shots requiring indirect personality expression (e.g., proxemics, spatial positioning, implicit behavioral cues) would yield lower personality alignment than shots with direct expression channels (e.g., facial expression, overt gestures). To test this, we compared PAS between direct and indirect shots using independent-samples Welch's $t$-tests for each model, pooling across all four conditions. Table~\ref{tab:indirect} reports the direct--indirect PAS comparison for each model.

\begin{table}[h]
\centering
\small
\begin{tabular}{@{}lcccc@{}}
\toprule
\textbf{Model} & \textbf{Direct} & \textbf{Indirect} & \textbf{$t$} & \textbf{$p$} \\
\midrule
SDXL & .430 & .423 & +0.24 & .812 \\
FLUX.1-dev & .497 & .467 & +1.15 & .253 \\
SD3.5-Large & .461 & .466 & $-$0.19 & .847 \\
StoryDiffusion & .477 & .484 & $-$0.29 & .770 \\
Qwen-2512 & .551 & .580 & $-$1.25 & .212 \\
GPT Image 2 & .590 & .601 & $-$0.32 & .749 \\
Nano Banana & .596 & .576 & +0.55 & .584 \\
\bottomrule
\end{tabular}
\caption{Direct vs.\ indirect PAS comparison (Welch's $t$-test). No model shows a significant difference ($p < .05$).}
\label{tab:indirect}
\end{table}

No model showed a significant difference between direct and indirect scenes (all $p > .20$). The mean PAS gap across seven models was $|\Delta| = 0.016$, with the direction inconsistent (three positive, four negative).

\paragraph{Interpretation.} This null result admits several explanations. First, current T2I models may not produce sufficiently fine-grained differences in indirect personality channels (e.g., interpersonal distance, spatial dominance) for VLM-based evaluation to detect. Second, the VLM judge itself may lack sensitivity to subtle proxemic and compositional cues, attending primarily to facial and gestural signals regardless of scene type. Third, the direct/indirect distinction may be less categorical than hypothesized: even ``indirect'' scenes often contain some facial or postural information that the VLM can exploit. Future work should design stronger indirect manipulations (e.g., scenes depicting only environmental or object-level personality cues without any character face visible) and employ more fine-grained evaluation metrics targeting specific visual channels.

\section{Per-Model Qualitative and Quantitative Analysis}
\label{app:per-model-analysis}
\begin{figure*}[t]
\centering
\setlength{\tabcolsep}{2pt}
\renewcommand{\arraystretch}{1.05}

\begin{tabular}{lcccc}
\toprule
\textbf{Model} & \textbf{C1} & \textbf{C2} & \textbf{C3} & \textbf{C4 / TBV} \\
\midrule

SDXL &
\includegraphics[width=0.18\textwidth]{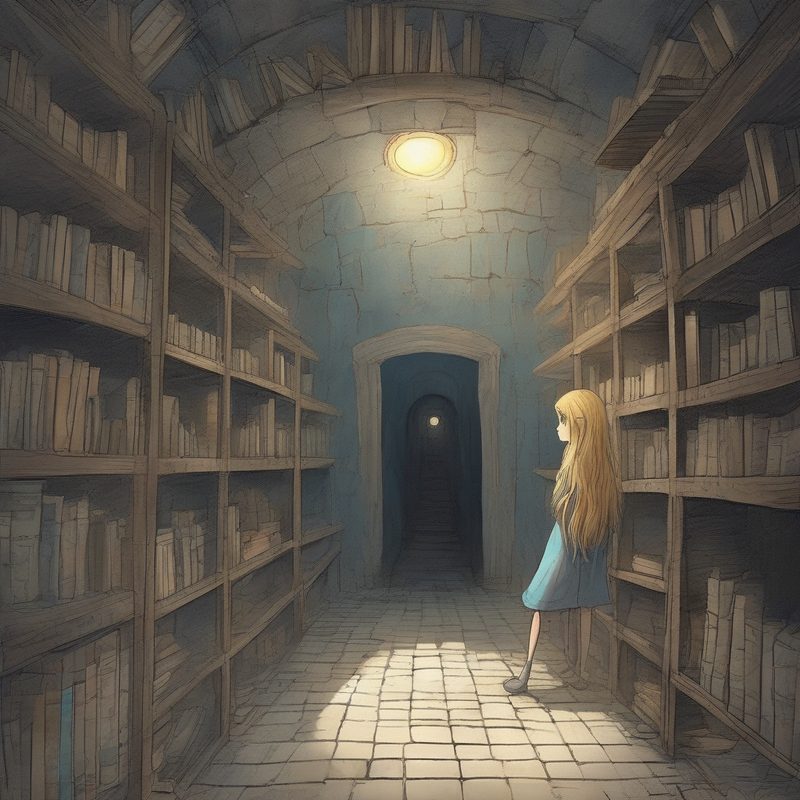} &
\includegraphics[width=0.18\textwidth]{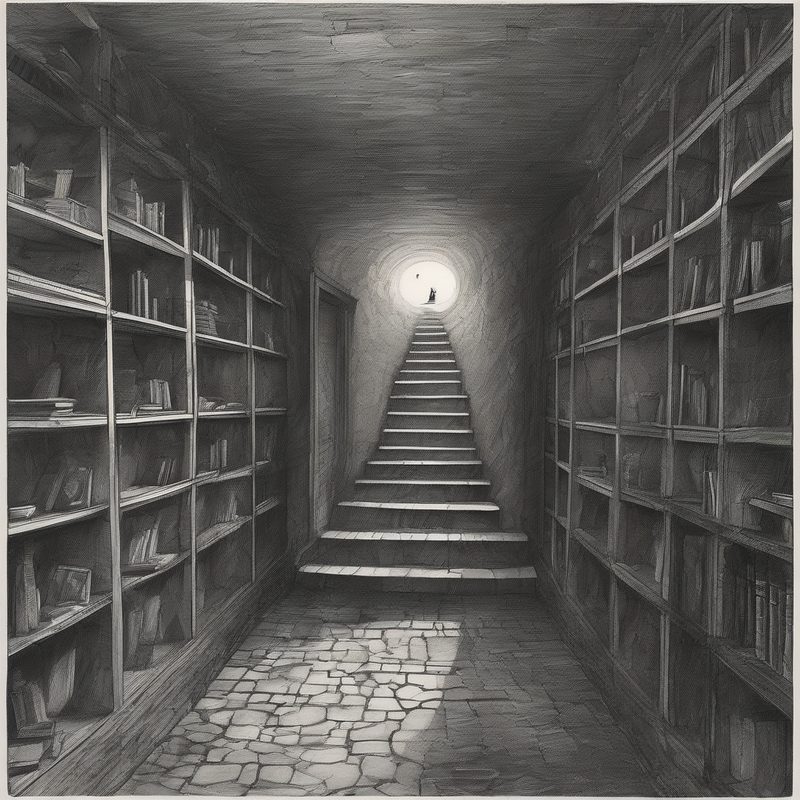} &
\includegraphics[width=0.18\textwidth]{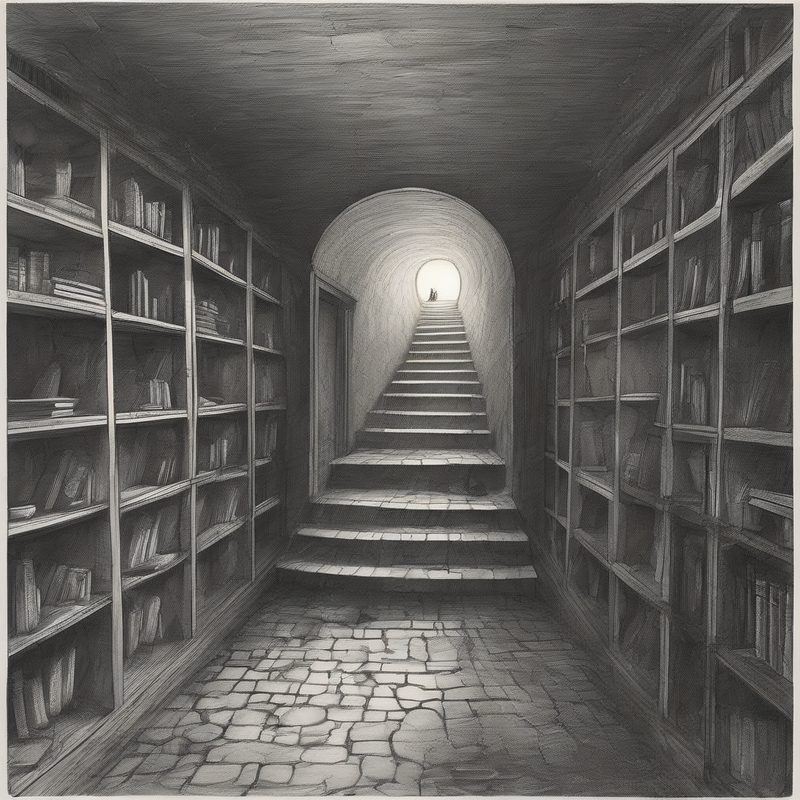} &
\includegraphics[width=0.18\textwidth]{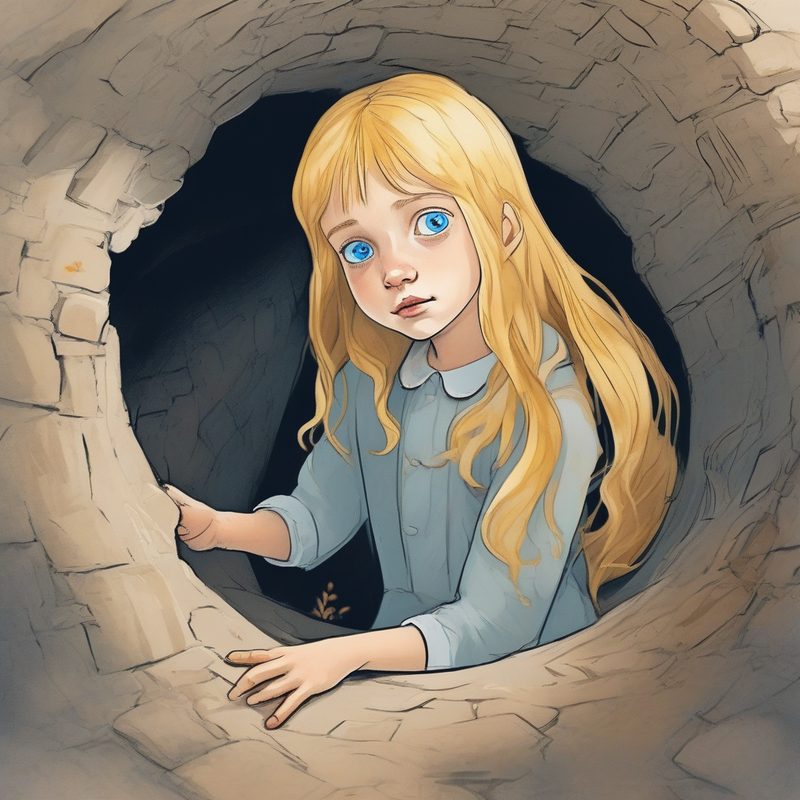} \\

StoryDiffusion &
\includegraphics[width=0.18\textwidth]{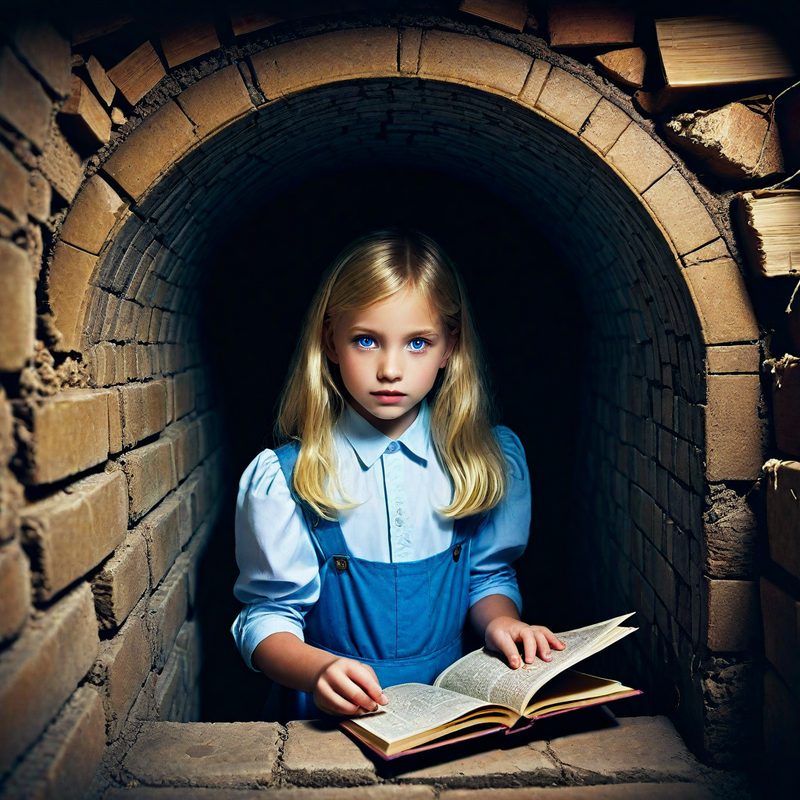} &
\includegraphics[width=0.18\textwidth]{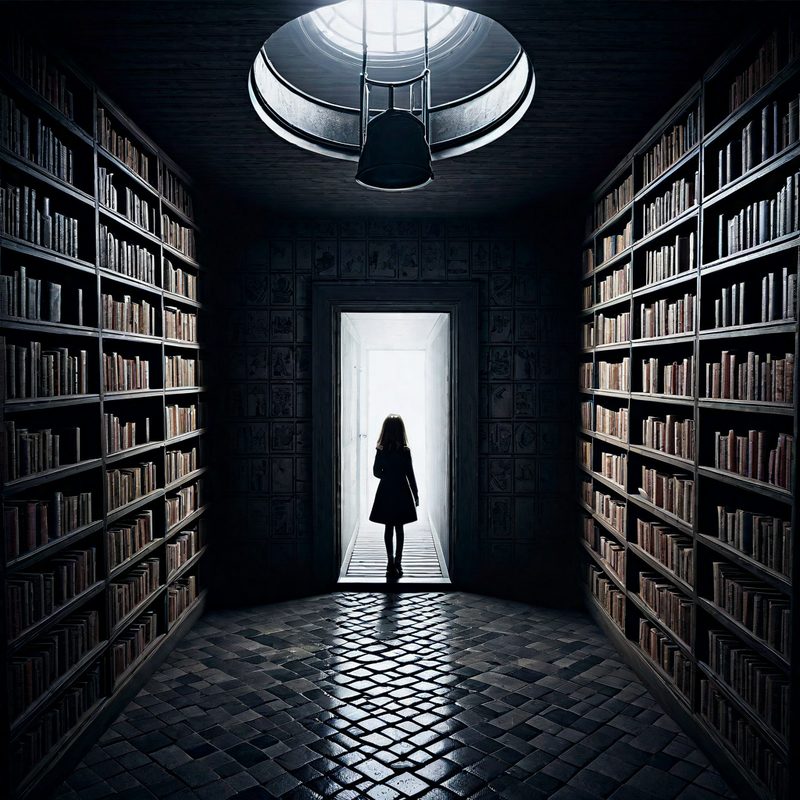} &
\includegraphics[width=0.18\textwidth]{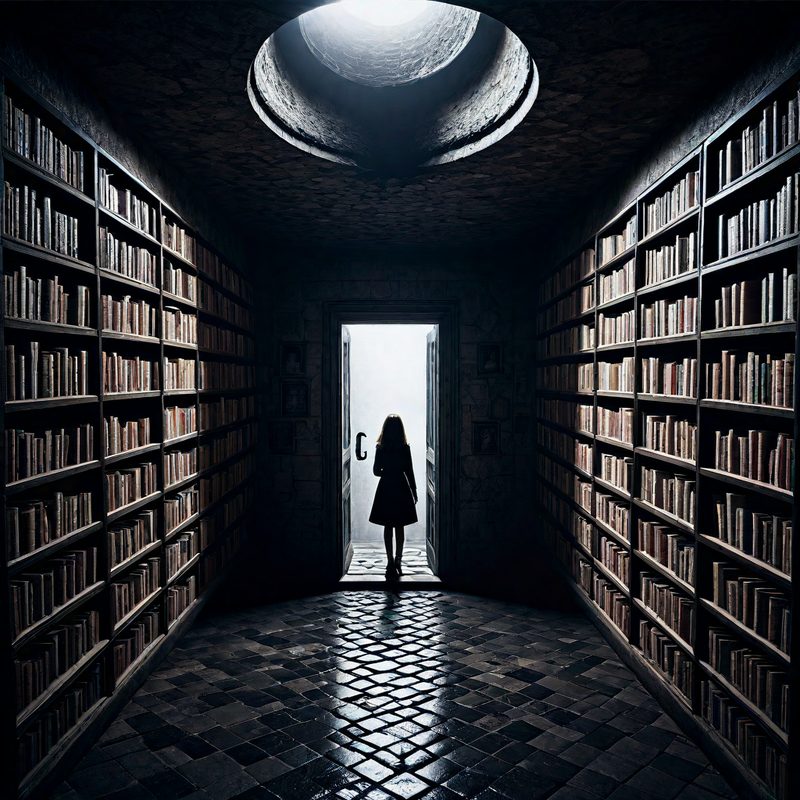} &
\includegraphics[width=0.18\textwidth]{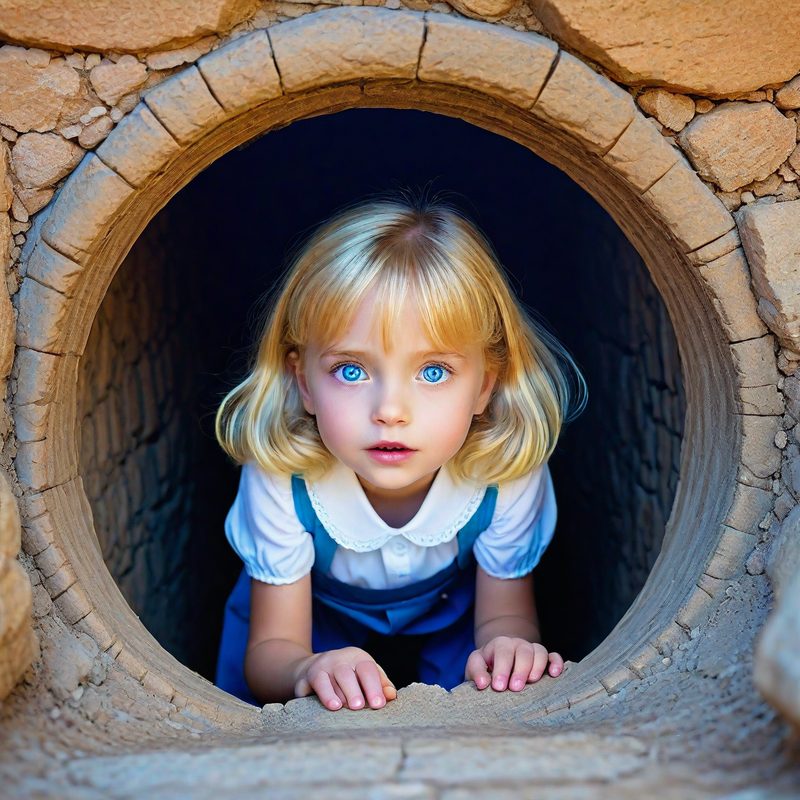} \\

FLUX.1-dev &
\includegraphics[width=0.18\textwidth]{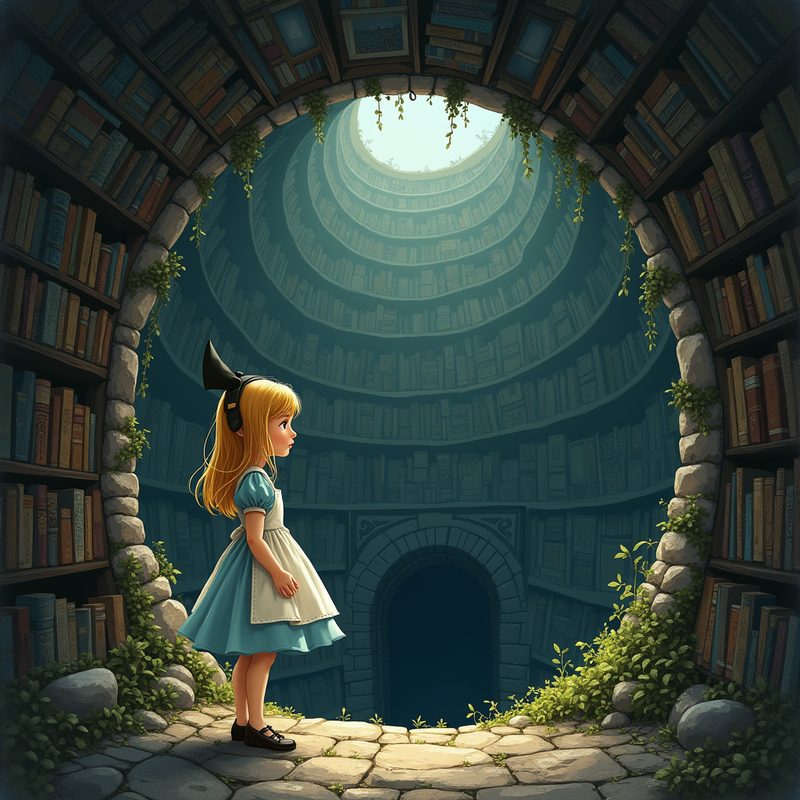} &
\includegraphics[width=0.18\textwidth]{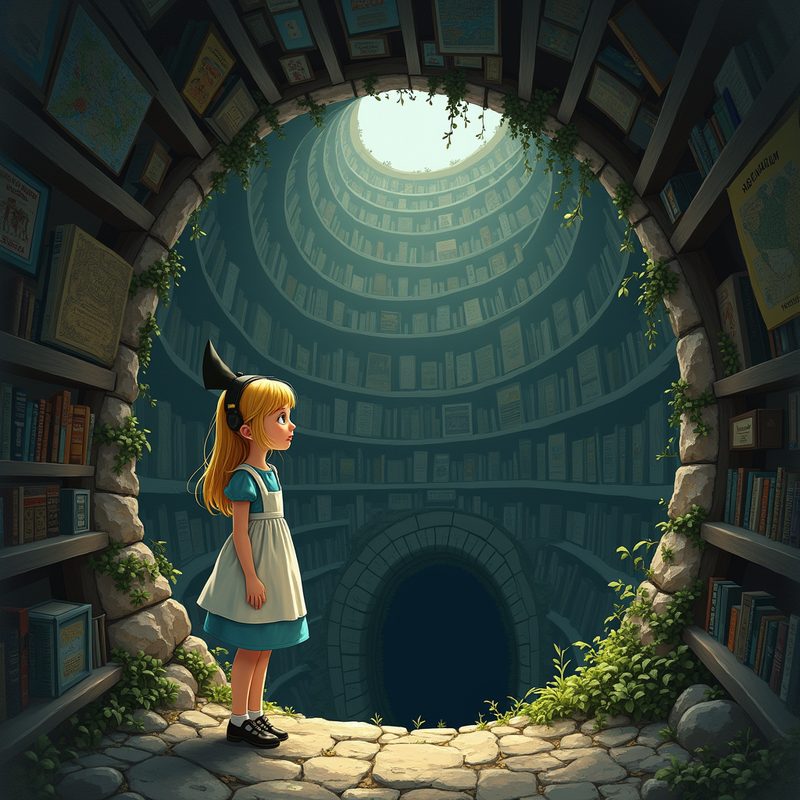} &
\includegraphics[width=0.18\textwidth]{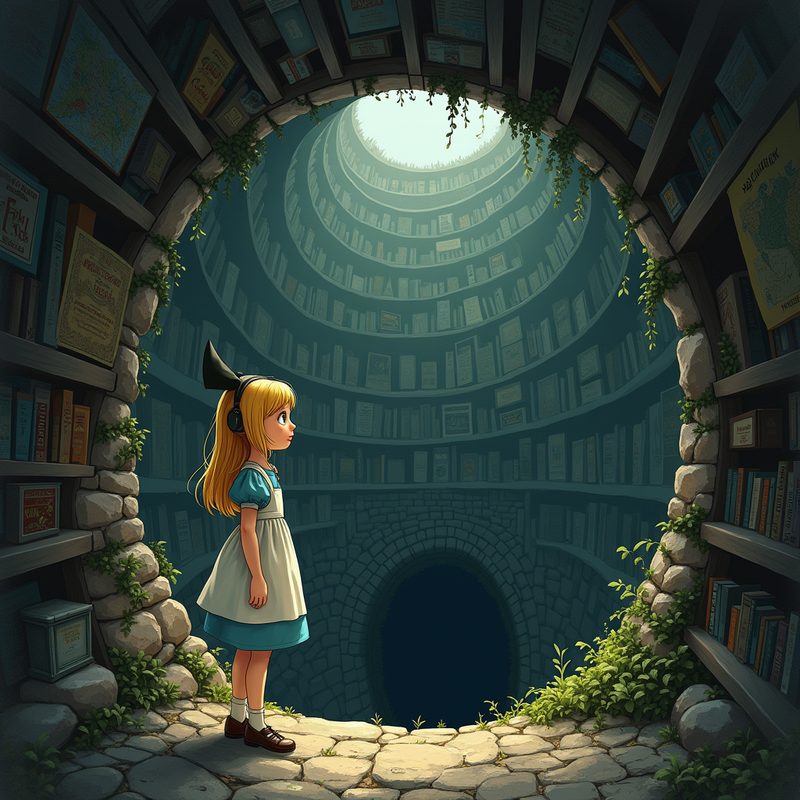} &
\includegraphics[width=0.18\textwidth]{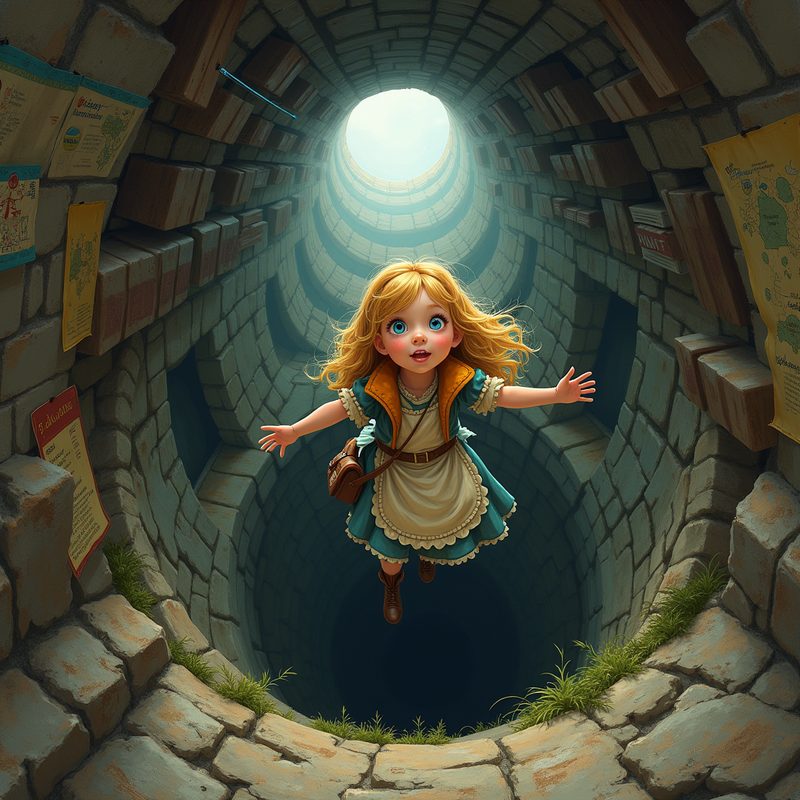} \\

SD3.5-Large &
\includegraphics[width=0.18\textwidth]{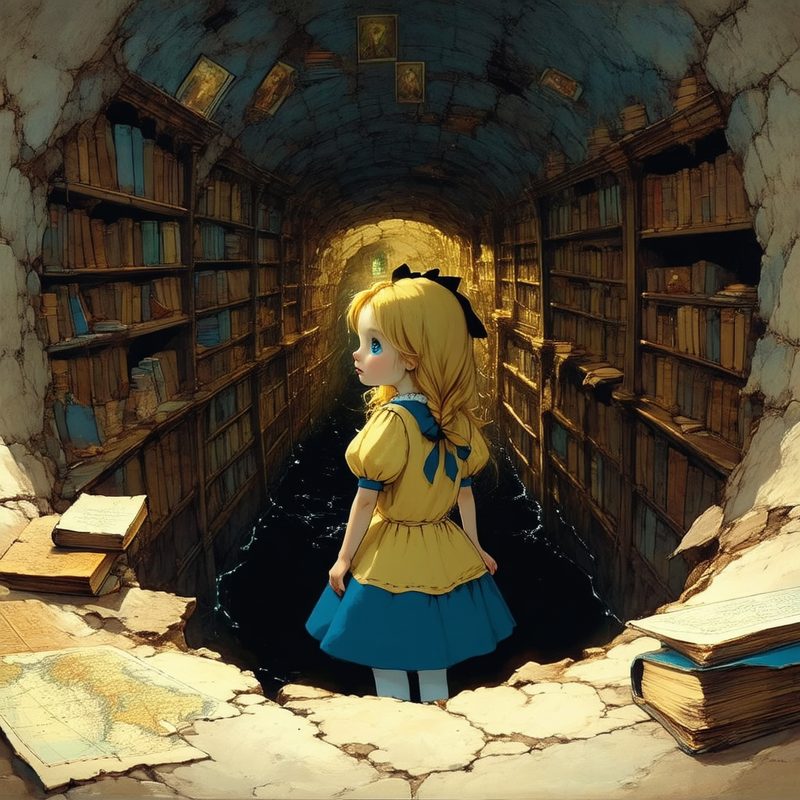} &
\includegraphics[width=0.18\textwidth]{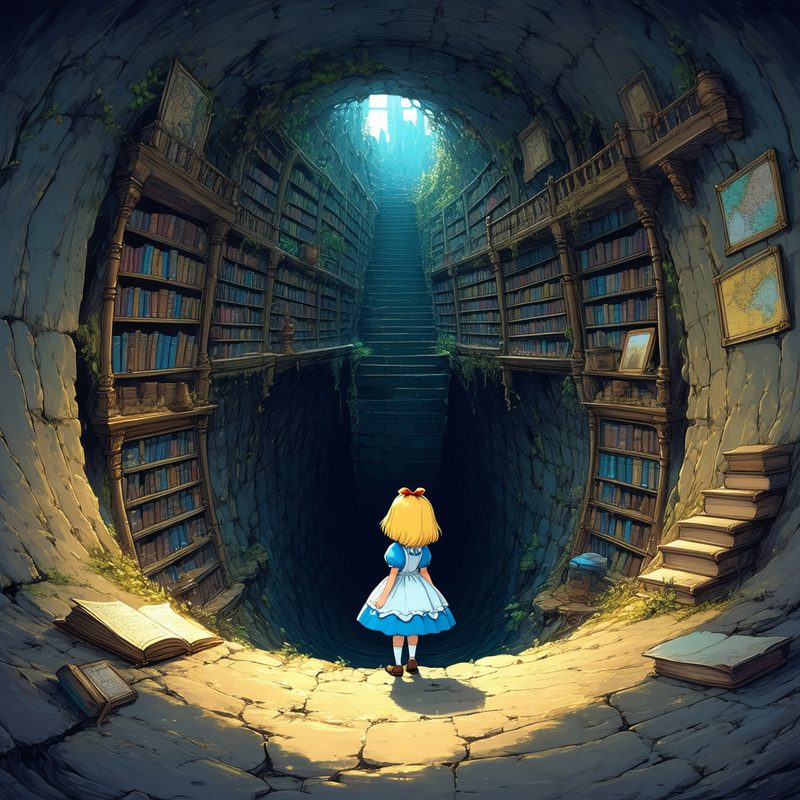} &
\includegraphics[width=0.18\textwidth]{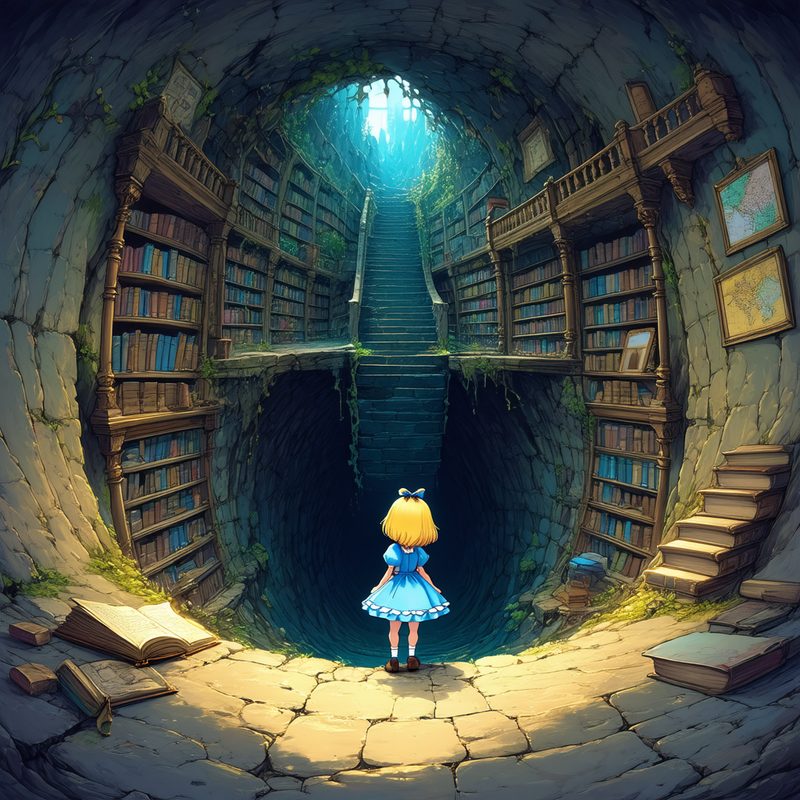} &
\includegraphics[width=0.18\textwidth]{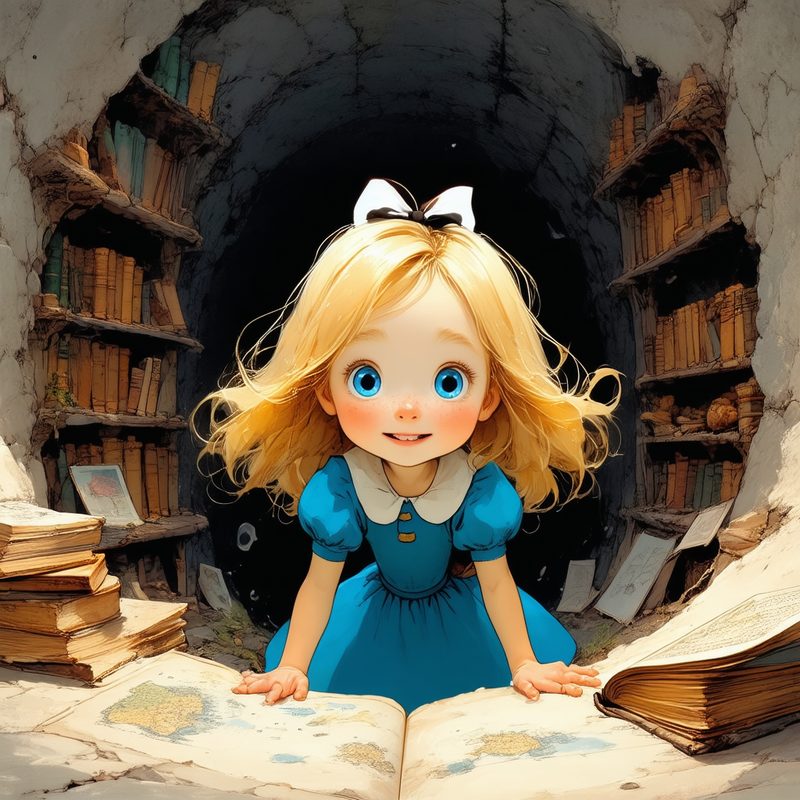} \\

Qwen-Image-2512 &
\includegraphics[width=0.18\textwidth]{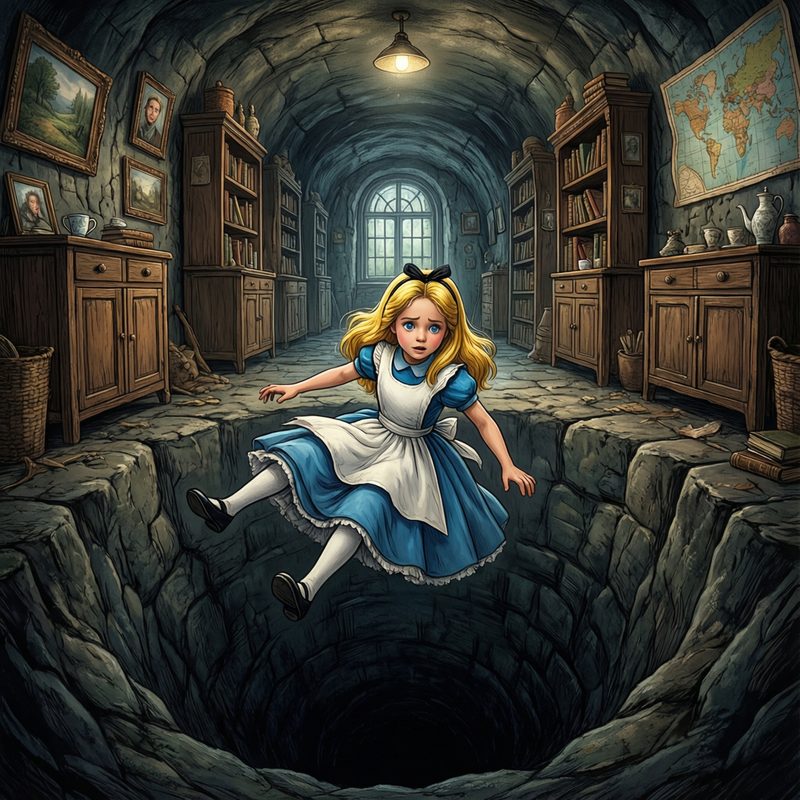} &
\includegraphics[width=0.18\textwidth]{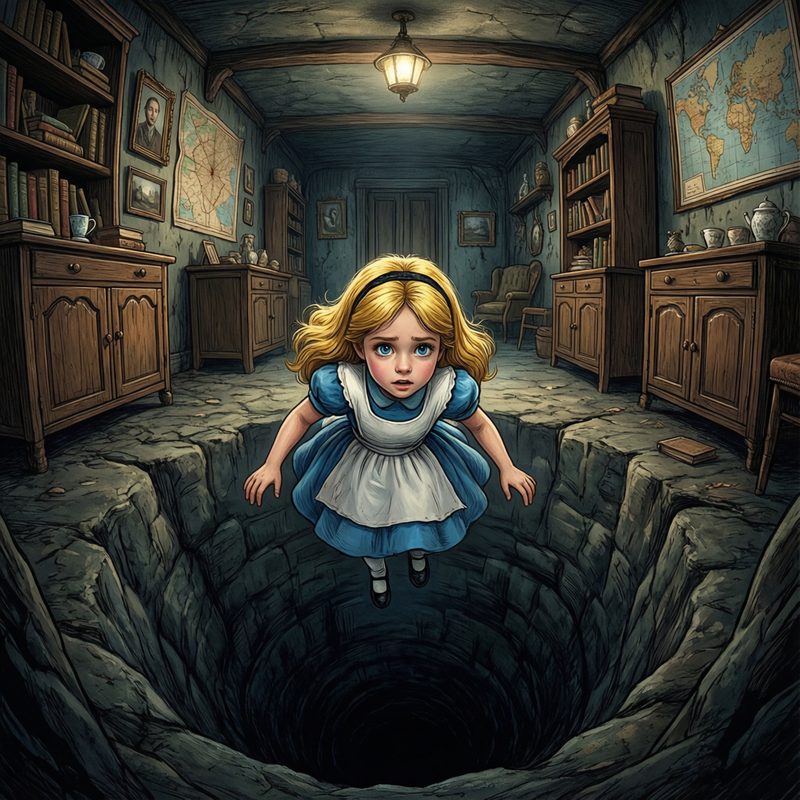} &
\includegraphics[width=0.18\textwidth]{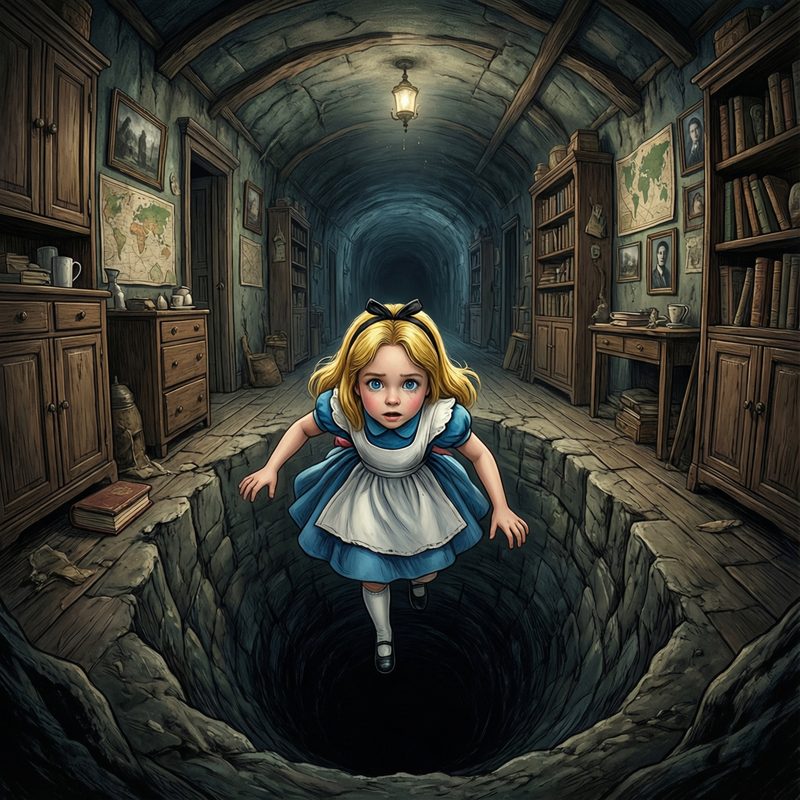} &
\includegraphics[width=0.18\textwidth]{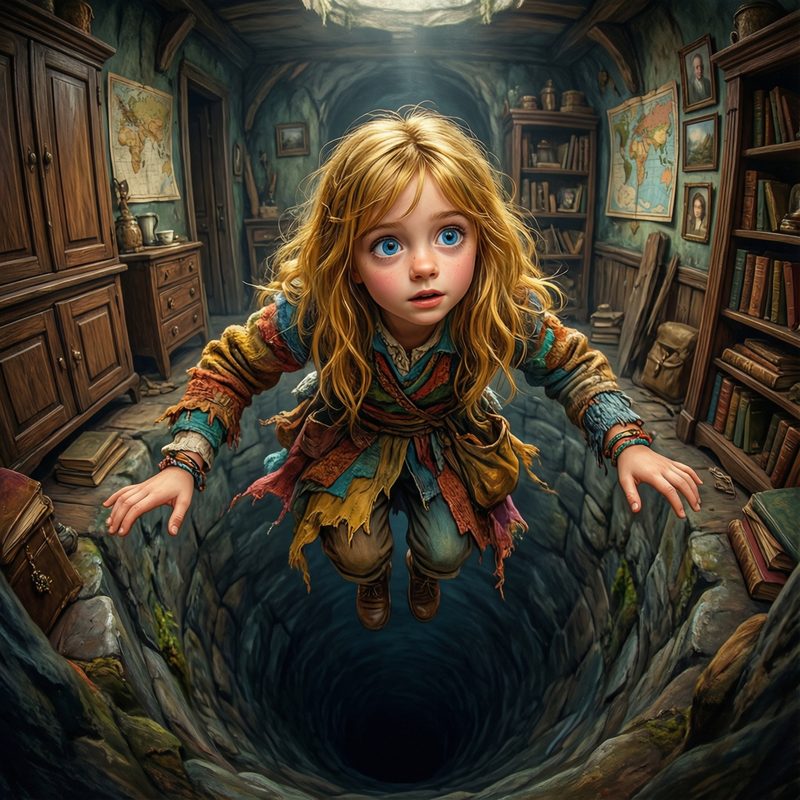} \\

\bottomrule
\end{tabular}

\caption{
Qualitative case study across five open-source systems on the same representative story shot. 
C1 uses only the base narrative prompt; C2 adds abstract OCEAN trait labels; C3 adds a deduced scene-level emotion label; C4 injects explicit TBV-derived visual constraints. 
Across models, C1--C3 often preserve the broad scene layout but tend to produce generic, weakly visible, or emotion-dominated behavior. 
C4 more consistently introduces personality-relevant visual evidence, such as exploratory gaze, forward-leaning posture, and open gestures, illustrating the qualitative mechanism behind the PAS and RS gains in Table~\ref{tab:main_results}.
}
\label{fig:qual_open}
\end{figure*}

\begin{figure*}[t]
\centering
\setlength{\tabcolsep}{4pt}
\renewcommand{\arraystretch}{1.05}

\begin{tabular}{lcc}
\toprule
\textbf{Model} & \textbf{C1} & \textbf{C4 / TBV} \\
\midrule

GPT Image 2 &
\includegraphics[width=0.32\textwidth]{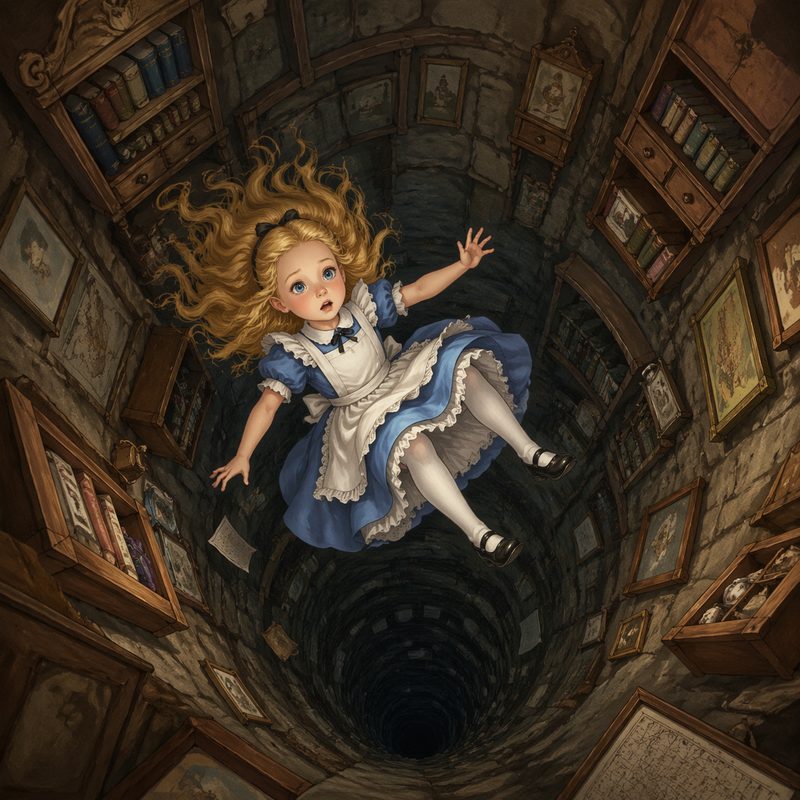} &
\includegraphics[width=0.32\textwidth]{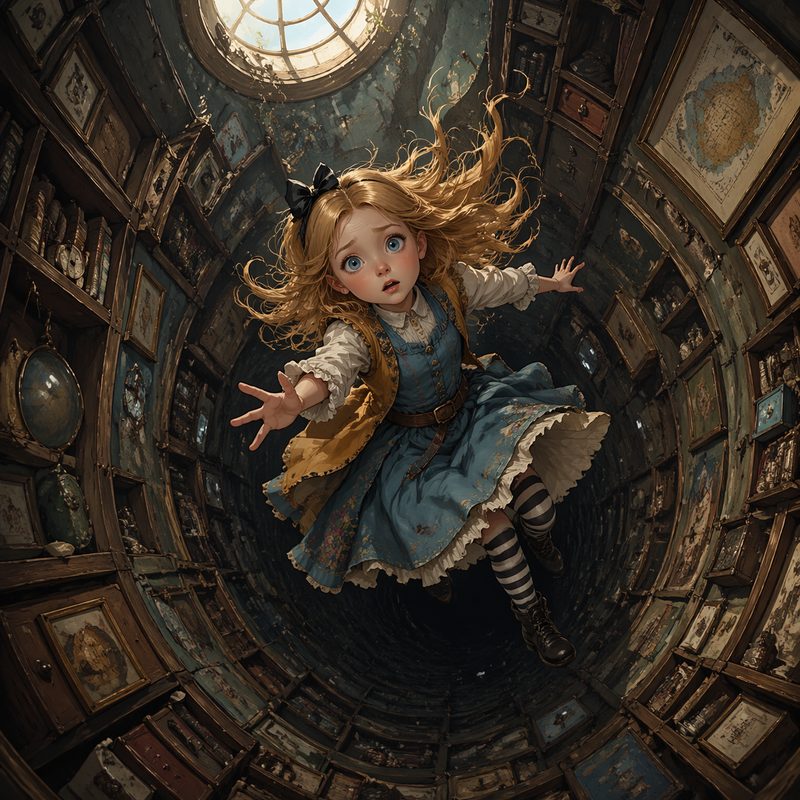} \\

Nano Banana &
\includegraphics[width=0.32\textwidth]{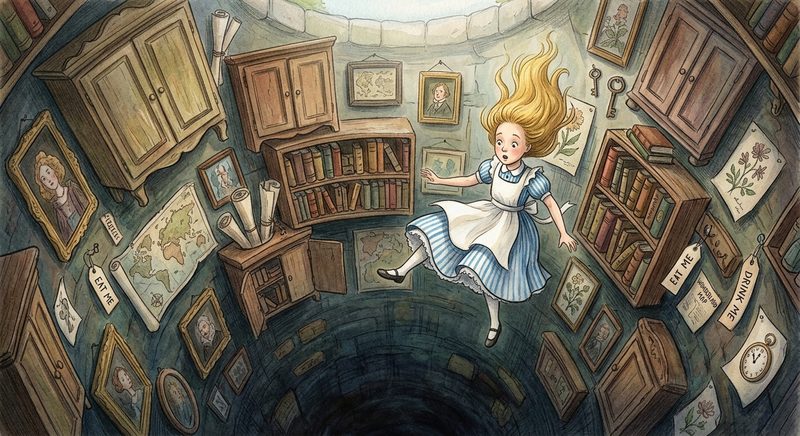} &
\includegraphics[width=0.32\textwidth]{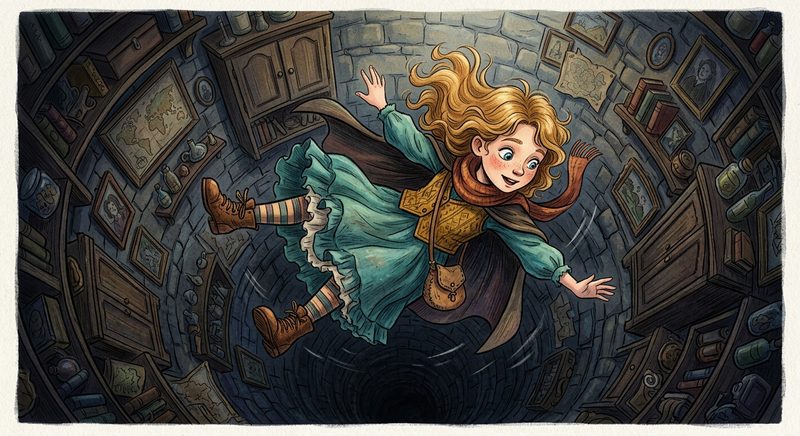} \\

\bottomrule
\end{tabular}

\caption{
Qualitative case study for the two closed-source APIs. 
Because GPT Image 2 and Nano Banana perform internal prompt rewriting, we evaluate them only under C1 and C4 conditions. 
Compared with C1, C4 produces more explicit personality-relevant behavior, including more active body orientation, exploratory engagement with the environment, and less purely scene-driven affect. 
This qualitative shift is consistent with the quantitative PAS and RS gains reported in Table~\ref{tab:main_results}.
}
\label{fig:qual_closed}
\end{figure*}

This appendix complements Table~\ref{tab:main_results} with model-wise qualitative analysis. 
Figures~\ref{fig:qual_open} and~\ref{fig:qual_closed} visualize the same representative story shot across systems: open-source models are compared under C1--C4, while closed-source APIs are compared only under C1 and C4 because they perform internal prompt rewriting.

Overall, C1--C3 usually preserve scene layout and character identity but often produce generic or emotion-dominated behavior. 
By contrast, C4 more often introduces TBV-specified cues such as exploratory gaze, forward-leaning posture, open gesture, and active environmental engagement. 
This qualitative shift mirrors the PAS and RS gains in Table~\ref{tab:main_results}, especially for models with stronger language-conditioning pathways.

\subsection{SDXL}

SDXL exhibits the weakest response to TBV among the evaluated systems. 
Its PAS increases only from 0.446 to 0.467, while RS slightly decreases from 0.429 to 0.423. 
This pattern suggests that CLIP-based text conditioning can partially absorb explicit visual cues, but remains limited in grounding abstract personality semantics into stable visual behavior. 
Qualitatively, SDXL often preserves the requested scene and character appearance, but the personality-relevant cues remain generic or emotionally dominated.

\subsection{StoryDiffusion}

StoryDiffusion provides a useful contrast between visual identity consistency and personality consistency. 
It achieves the strongest baseline PCS among open-source systems, but its baseline PAS remains moderate. 
This indicates that consistent self-attention improves cross-frame visual stability without necessarily preserving personality-relevant behavior. 
After TBV prompting, PAS improves from 0.480 to 0.502, but the gain remains smaller than models with stronger language understanding. 
This supports our claim that personality grounding is not solved by visual consistency mechanisms alone.

\subsection{FLUX.1-dev}

FLUX.1-dev shows a stronger response to TBV than SDXL and StoryDiffusion. 
PAS increases from 0.464 to 0.522, and RS increases from 0.469 to 0.552. 
This suggests that FLUX can better exploit concrete visual constraints once personality is translated into renderable cues. 
Qualitatively, TBV prompts tend to produce clearer posture, gaze, and interaction cues, although the model still occasionally prioritizes scene-level affect over stable personality expression.

\subsection{SD3.5-Large}

SD3.5-Large shows only marginal improvement under TBV, with PAS increasing from 0.466 to 0.472 and RS slightly decreasing from 0.493 to 0.488. 
This indicates that a stronger generative architecture does not automatically yield stronger personality grounding. 
One possible explanation is that the model may normalize or weaken fine-grained prompt details during generation, reducing the effect of additional visual constraints. 
Its results suggest that architectural sophistication alone is insufficient unless the text-conditioning pathway preserves personality-relevant behavioral cues.

\subsection{Qwen-Image-2512}

Qwen-Image-2512 is the strongest open-source responder to TBV. 
PAS increases from 0.518 to 0.633, RS from 0.546 to 0.641, and SCS from 0.665 to 0.701. 
This model benefits most from the TBV decomposition, suggesting that LLM-conditioned T2I systems are better able to use explicit visual constraints derived from abstract traits. 
Qualitatively, Qwen-Image-2512 more reliably renders exploratory gaze, forward-leaning posture, and spontaneous gestures when these cues are explicitly specified.

\subsection{GPT Image 2}

GPT Image 2 shows the strongest improvement among closed-source systems. 
PAS increases from 0.538 to 0.648, while RS increases from 0.556 to 0.659. 
This result supports the broader trend that models with stronger language understanding are more responsive to TBV prompting. 
However, because GPT Image 2 performs internal prompt rewriting, we evaluate it only under C1 and C4 conditions and do not use it for full C1--C4 ablation analysis.

\subsection{Nano Banana}

Nano Banana starts from a relatively strong baseline and still benefits substantially from TBV. 
PAS increases from 0.542 to 0.638, and RS increases from 0.551 to 0.661. 
PCS slightly decreases, suggesting that TBV may introduce more personality-specific variation across scenes while improving snapshot-level personality alignment. 
This trade-off reinforces the need to evaluate both alignment and stability rather than relying on a single consistency metric.

\subsection{Summary}

Across models, the largest TBV gains occur in LLM-conditioned or stronger language-conditioned systems, while CLIP-only or visual-consistency-focused systems show smaller gains. 
These model-wise patterns support two conclusions. 
First, visual identity preservation does not imply personality preservation. 
Second, the key bottleneck in trait collapse lies primarily in semantic grounding: models require concrete visual constraints before abstract personality traits become reliably renderable.

\end{document}